%% file: main.tex
\documentclass[runningheads]{llncs}

\usepackage{eccv} 

\usepackage{eccvabbrv}
\usepackage{graphicx}
\usepackage{subcaption} 
\usepackage{booktabs}
\usepackage{multirow} 
\usepackage{amsmath}
\usepackage{amssymb}
\usepackage[accsupp]{axessibility}
\usepackage[most]{tcolorbox}
\usepackage{tikz}
\usetikzlibrary{arrows.meta, fit, positioning, backgrounds}
\usepackage{wrapfig,lipsum}
\usepackage[table]{xcolor}   
\usepackage{placeins} 
\usepackage{bibunits}
\usepackage{setspace}
\usepackage{enumitem}
\usepackage{xcolor,soul}
\usepackage{orcidlink}
\usepackage{hyperref}

\newcommand{\auc}{\mathrm{AUC}}
\newcommand{\AUCglobal}{\auc_{\mathrm{global}}}
\newcommand{\AUCcat}{\auc_{\mathrm{cat}}}
\newcommand{\AUCwv}{\auc_{\mathrm{wv}}}

\setlist[enumerate]{nosep,leftmargin=1.4em,label=\alph*}

\definecolor{termcol}{RGB}{178,34,34}          
\definecolor{termhl}{RGB}{173,216,230}   
\sethlcolor{termhl}

\definecolor{takeawayblue}{RGB}{75,111,165}
\definecolor{takeawayfill}{RGB}{225,232,243}
\definecolor{tan}{RGB}{247,223,178}  
\definecolor{green}{RGB}{201,231,192}  
\definecolor{blue}{RGB}{188,212,230}     
\definecolor{lightred}{RGB}{253,213,213} 
\definecolor{objectionred}{RGB}{170,45,45}
\definecolor{objectionfill}{RGB}{252,235,235}
\definecolor{cbar}{RGB}{0,114,178}

\definecolor{rankone}{RGB}{190,240,190}   
\definecolor{ranktwo}{RGB}{213,247,213}   
\definecolor{rankthree}{RGB}{233,251,233}  

\newcommand{\rkone}[1]{\cellcolor{rankone}\textbf{#1}}
\newcommand{\rktwo}[1]{\cellcolor{ranktwo}#1}
\newcommand{\rkthree}[1]{\cellcolor{rankthree}#1}

\newif\ifcorrectionsfinal
\correctionsfinaltrue
\definecolor{corrbg}{RGB}{255,213,145}
\ifcorrectionsfinal
  \newcommand{\corr}[1]{#1}
\else
  \newcommand{\corr}[1]{\colorbox{corrbg}{\strut #1}}
\fi
\ifcorrectionsfinal
  \newcommand{\corrpar}[1]{#1}
\else
  \sethlcolor{corrbg}
  \newcommand{\corrpar}[1]{\hl{#1}}
\fi

\begin{document}
\begin{bibunit}[splncs04]

\title{Auditing Frame-Level AUC in Weakly Supervised Video Anomaly Detection:\\ Granularity, Resolution, and Scene Bias}
\titlerunning{Auditing Frame-Level AUC in WSVAD} 

\author{Sara Abdulaziz\inst{1}\orcidlink{0009-0002-5204-127X} \and Egor Bondarev \inst{1}\orcidlink{0009-0005-2452-7389} }
\authorrunning{S. Abdulaziz and E. Bondarev}

\institute{Eindhoven University of Technology, 5612 AE Eindhoven, Netherlands\\ \email{s.e.a.m.abdulaziz@tue.nl}\\
\href{https://github.com/Sara-Esam/vad_auc_audit}{GitHub Code}
}

\maketitle

\begin{abstract}
Frame-level area under the ROC curve (AUC) is the dominant evaluation metric for weakly supervised video anomaly detection (WSVAD). Its standard form measures whether an anomalous frame outranks a normal frame drawn from anywhere in the test set. We refer to this comparison as \emph{pooled AUC}, since it aggregates frame pairs across test videos regardless of source. Pooled AUC therefore credits both event localization and differences between video sources. We audit this protocol on UCF-Crime~\cite{sultani2018real} across recent state-of-the-art models spanning different backbone families. Holding each model's frame scores fixed, we read them under three pairing granularities: global, per anomaly category, and within each video, then repeat the same three-granularity readout on zero-shot scores computed from the models' internal representations. We assess ranking reliability with a paired video bootstrap. Three findings follow. First, pooled AUC does not reliably predict within-video anomaly localization: models with similar pooled scores exhibit large localization differences and rank reversals under stricter granularities. Second, at the benchmark's test-split size, pooled AUC lacks the resolution to support state-of-the-art margins reported in the field. Within each evaluated backbone family, it resolves no comparison at those margins, while within-video AUC resolves several over identical predictions. Learned representations further reveal that within-video anomaly structure and detector localization are decoupled. Third, on normal footage alone, every model we examine separates videos by recording properties, such as resolution and color encoding, indicating that scene sensitivity is shared across the setting rather than specific to any architecture. We publicly release a granularity-aware protocol computable from existing predictions and scene-factor annotations for UCF-Crime.

\end{abstract}

\section{Introduction}
\label{sec:intro}

Weakly supervised video anomaly detection (WSVAD) aims to localize anomalous events in long videos while using only video-level labels during training~\cite{sultani2018real,urdmu,wu2024vadclip}. Progress on VAD benchmarks~\cite{sultani2018real,zhu2024advancing} is commonly summarized by a single metric: frame-level anomaly prediction measured by the ROC-AUC (area under the true positive rate vs.\ false positive rate curve), pooled across all anomalous-normal frame pairs in the test set. AUC is threshold independent and admits a useful ranking interpretation statistically. However, its statistical consistency concerns \emph{precision} at a given sample size, and does not guarantee that the \emph{target objective} is being measured~\cite{vskvara2023auc,li2024area,mcdermott2024closer}. A consistent estimator may still be too imprecise to rank at a finite sample $n$, or can be precise but estimating the wrong quantity.

A WSVAD model scores positively in an AUC comparison whenever an anomalous frame receives a larger anomaly score than a normal frame, regardless of whether the ordering follows the event semantics, behavior, or the footage source. This exposes two distinct failures, indistinguishable by the metric. First, a high AUC may not imply genuine anomaly detection. As the credited comparisons are overwhelmingly cross-video, a model's predictions can be rewarded by AUC while learning to separate recording scenes rather than actual anomaly events. At deployment, a detector must rank anomalous intervals above normal ones from the same camera stream. Thus, a model relying on cross-video separation might produce constantly elevated scores, permanent false alarm, on anomaly-associated cameras while missing anomalies in atypical contexts. Pooled AUC cannot distinguish such failure from true localization. Second, the cross-video evaluation protocol yields an AUC measure that is too coarse to support model ranking reliably. As the easy cross-video pairs dominate the pool, they compress most models AUC toward a common narrow range, which may fall within the metric's own sampling error~\cite{hanley1982meaning, delong1988comparing}.

In this work, we investigate the evaluation process of WSVAD models. Using the same frame-level predictions produced by existing methods, we ask: (i) what does the standard pooled evaluation actually measure, and does it resolve models ranking reliably? (ii) Does the anomaly signal dominate in each model's learned representation, and to what extent do the final anomaly scores depend on it? (iii) What determines anomaly score responses in entirely normal video segments? (iv) Does the pooled AUC ranking hold when comparisons are restricted to frames from the same recording? To answer these questions, we make the following contributions:

\begin{itemize}
    \item A multi-granularity evaluation framework that reads the same predictions under different pairing rules, disentangling cross-recording based separation from within-video anomaly localization. We apply this to both final scores and training-free prototypes constructed from learned representations.
    \item A resolution analysis of WSVAD benchmarks: by resampling test videos, we assess which pairwise model comparisons are statistically distinguishable at each granularity. We reveal that standard pooled AUC cannot separate models at the sub-point margins the field reports, whereas within-video AUC can.
    \item A scene-reliance audit of over twelve annotated binary scene factors, computed on normal frames only. We show that every examined model separates frames by recording properties in the absence of anomalies.
    \item We publicly release (i) the scene-factor annotations for the full UCF-Crime dataset, and (ii) code for an evaluation protocol computable from existing predictions.
\end{itemize}

\section{Related Work}
\label{sec:related}
\paragraph{Weakly supervised video anomaly detection.}
Since the introduction of the large-scale UCF-Crime benchmark~\cite{sultani2018real}, weakly supervised video anomaly detection has evolved rapidly. Models have progressed from basic temporal instance selection to increasingly sophisticated mechanisms, including robust normality modeling, memory-augmented representations, and multimodal vision-language alignments~\cite{urdmu,zhou2024batchnorm,pel4vad,pivad,gs-moe,wu2024vadclip,dsanet}. Across all works, methods are still predominantly ranked by pooled AUC on UCF-Crime, which is the protocol under question in this study. Rather than proposing a new VAD model, we examine whether this evaluation protocol supports the temporal-localization conclusions drawn from the same models and predictions. 

\paragraph{Evaluation protocols for video anomaly detection.}
Limitations of frame-level aggregation have previously been identified from several complementary perspectives. Ramachandra and Jones~\cite{ramachandra2020street} introduced region-based and track-based detection criteria for anomalous object and trajectory localization. This resolves the limitation of the frame-level AUC, which rewards a video frame whenever any score exceeds the threshold regardless of the event continuity. Their analysis exposes a spatial credit-assignment problem, where the rank statistic may be computed correctly while the set of credited detections does not correspond to the intended task. Liu \etal~\cite{liu2025rethinking} identify three limitations of current VAD evaluation, which are the reliance on a single-perspective temporal annotations, failure to reward early anomaly detection, and the inability of existing benchmarks to reveal scene overfitting. The authors address these limitations with annotation-averaged AUC/AP, Latency-aware Average Precision (LaAP), and hard-normal benchmarks constructed to test whether models generalize beyond anomaly-associated scenes. Rashvand \etal~\cite{rashvand2026frames} focus on the limitation of frame-level evaluation, where it treats a video as a collection of isolated frames, while an anomaly is a coherent temporal event with an onset and duration. The authors introduce an event-centric protocol based on temporal-IoU event matching and multi-threshold event-level \(F_1\).

\paragraph{AUC and VAD model comparison.}

AUC estimates the probability that a randomly selected anomalous segment receives a higher score than a randomly selected normal segment. Even though its threshold independence and relative stability are useful for binary frame classification, pooled AUC can obscure the operating region and anomaly distribution of practical interest~\cite{vskvara2023auc}. Defenses of AUC often conflate two distinct properties: (1) the \textit{prevalence invariance} of the estimator, which characterizes its stability across varying positive-negative ratios within the estimand~\cite{li2024area,mcdermott2024closer}, and (2) the \emph{sampling precision} of the estimator at finite test size $n$. Only the former is granted by the AUC's ranking interpretation. The latter is quantified analytically by Hanley and McNeil~\cite{hanley1982meaning,mcneil1984statistical}, showing that the sampling standard errors of AUC scale by $\frac{1}{\sqrt{n}}$ in the effective sample size, regardless of the positive-negative sample balance. The effective samples size in temporally correlated data, such as video frames, is closer to the number of videos than to the number of frames. At the scale of current WSVAD test sets~\cite{sultani2018real}, the standard sampling error might be order of magnitudes larger than the margins by which SoTA is currently claimed. Our video-level bootstrap verifies this empirically.

\paragraph{Benchmark, scene, and representation auditing.}
A parallel line of work in the context of evaluation asks whether benchmark performance reflects the intended evidence rather than dataset-specific shortcuts~\cite{eyuboglu2022domino,d2022spotlight,ribeiro2020beyond,zverev2025vggsounder,mumcu2026video}. 
For example, Mumcu \etal~\cite{mumcu2026video} question whether multi-scene anomaly recognition is appropriately formulated, arguing that scene and anomaly semantics can become entangled under current benchmarks. This concern motivates our scene-reliance audit, but our analysis is performed at the evaluation level. 

\section{Granularity-aware VAD Evaluation}
\label{sec:method}

\begin{figure}[t]
    \centering
    \includegraphics[width=0.70\linewidth]{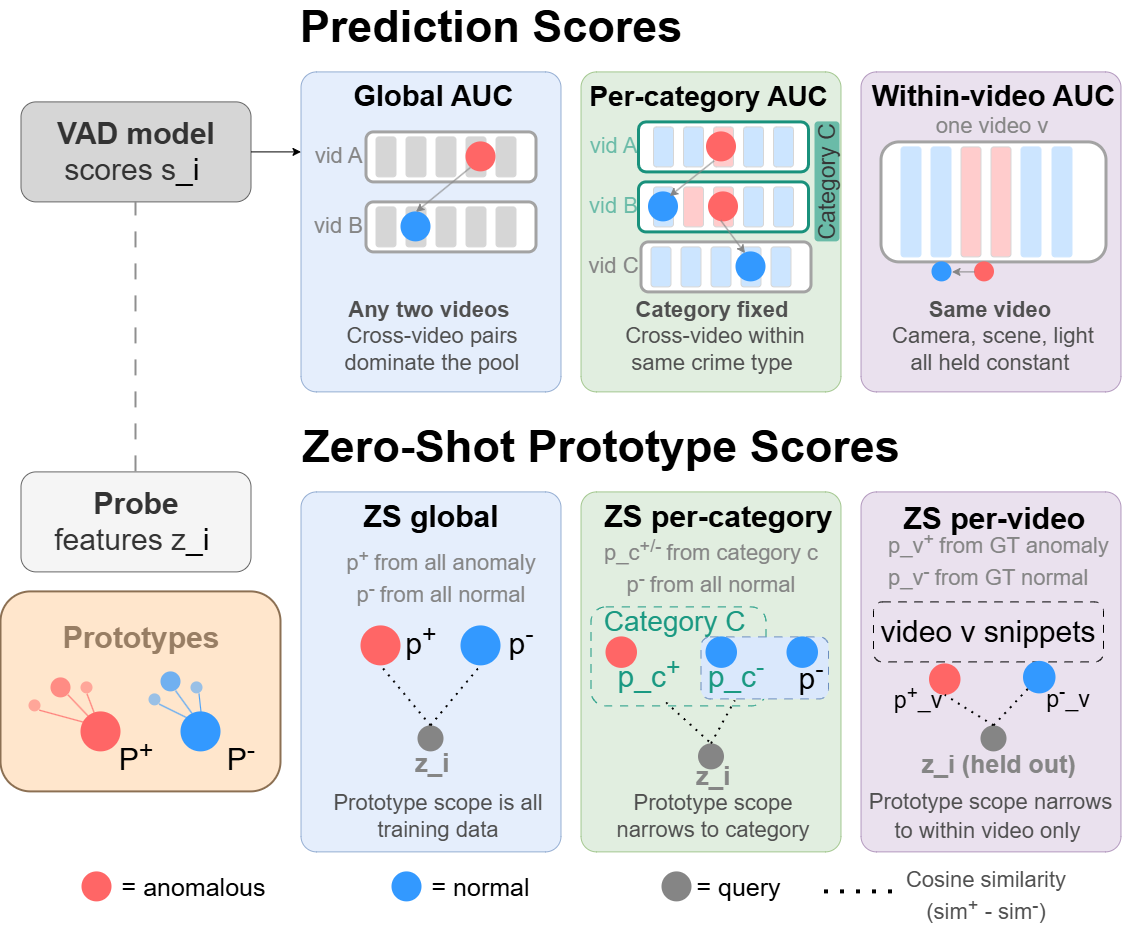}
    \caption{\textbf{Two score sources for AUC under three pairing rules.} Each scoring mechanism ranks anomalous (red) against normal (blue) frames, differing only in which pairs are eligible. Top: the detector's own prediction scores $s_i$. Bottom: prototype scores, where the pre-classification head embeddings from frozen VAD models are applied to construct a normal and abnormal class-mean (prototype). Then, an anomaly score for a snippet feature $z_i$ is computed by its similarity to anomalous versus normal prototypes. 
    }
    \label{fig:three-granularities-fig}
\end{figure}

Let $s_i \in \mathbb{R}$ be the anomaly score of frame $i$, $y_i \in \{0,1\}$ its frame-level label, $v_i$ its video identity, and $c_i$ the anomaly category of its video. We compute AUC at three granularities by changing: (1) the set of eligible positive-negative frame pairs, and (2) the anomaly score source.~\Cref{fig:three-granularities-fig} demonstrates this evaluation protocol.

\subsection{Prediction-Score Granularities}
We evaluate a model's frame-level scores under three pairing rules for AUC that differ only in which anomalous-normal frame pairs are eligible for comparison: \emph{global} (any two videos), \emph{per-category} (same anomaly category videos), and \emph{within-video} (same video). For all, models scores are fixed and we compute the same ROC-AUC statistic varying only the population. 

\paragraph{Global AUC.}
The standard frame-level protocol established in the WSVAD field~\cite{sultani2018real}, which pools the full test set frames:
\begin{equation}
\AUCglobal
=
\Pr\!\left(s^{+} > s^{-}\right),
\label{eq:auc-global}
\end{equation}
where $\Pr(\cdot)$ denotes probability under a uniform random draw of one anomalous frame score $s^{+}$ and one normal frame score $s^{-}$ from the frame pool in the test set. Since the anomalous and normal frames may originate from any video; behavioral and contextual separability are both rewarded in $\AUCglobal$. We use \emph{global} and \emph{pooled} interchangeably for the standard protocol.

\paragraph{Per-category AUC.}
For anomaly category $c$, where $c\in\mathcal{C}$, positives are the anomalous frames of videos labeled $c$, and negatives are the normal frames of those same videos. Normal frames inherit the category of the video they occur in, and videos originally annotated as normal do not contribute at this granularity. A single AUC is computed over each category's full frame pool. The reported value averages these across categories,
\begin{equation}
\AUCcat
=
\frac{1}{|\mathcal{C}|}
\sum_{c\in\mathcal{C}}
\auc\!\left(
\{s_i : y_i = 1,\, c_i = c\},
\{s_i : y_i = 0,\, c_i = c\}
\right),
\label{eq:auc-category}
\end{equation}
such that each category contributes equally regardless of the number of videos or frames it contains. Fixing the category constrains the type of event that the positives depict, but leaves the recording context free. Specifically, an anomalous frame from one video is still ranked against a normal frame from another. This granularity is therefore stricter than global pooling with respect to category composition, while still admitting cross-video
comparisons.

\paragraph{Within-video AUC.}
Let $\mathcal{V}_{\pm}$ contain test videos with at least one anomalous and one normal frame. We compute an AUC inside every video and macro-average across videos:
\begin{equation}
\AUCwv
=
\frac{1}{|\mathcal{V}_{\pm}|}
\sum_{v\in\mathcal{V}_{\pm}}
\auc\!\left(
\{s_i:y_i=1,v_i=v\},
\{s_i:y_i=0,v_i=v\}
\right).
\label{eq:auc-within-video}
\end{equation}
The compared frames share the same recording, camera, location, and broad acquisition conditions. The model must rank the anomalous interval above the video's own normal footage. We therefore interpret $\AUCwv$ as \textit{a temporal localization measure}, not as a replacement for global detection performance.

\subsection{Zero-Shot Representation Readout}
\label{sec:zs}

Prediction scores characterize the final WSVAD model but cannot identify whether a failure originates in the representation or the scoring head. We therefore repeat the same three-granularity readout using zero-shot prototype scores on the internal features of WSVAD models, extracted immediately before the models final classification head.

Let $z_i\in\mathbb{R}^{d}$ represent the feature vector extracted from WSVAD model for snippet $i$ before the classification head. For a given granularity level $g$, denote the anomalous and normal support sets by $\mathcal{S}_{g}^{+}$ and $\mathcal{S}_{g}^{-}$, respectively. Their corresponding prototypes $p_{g}^{\pm}$ are defined as the mean of feature vectors in each class

\begin{equation}
p_{g}^{+}
=
\frac{1}{|\mathcal{S}_{g}^{+}|}
\sum_{j\in\mathcal{S}_{g}^{+}} z_j,
\qquad
p_{g}^{-}
=
\frac{1}{|\mathcal{S}_{g}^{-}|}
\sum_{j\in\mathcal{S}_{g}^{-}} z_j,
\label{eq:zs-prototypes}
\end{equation}

where $p_g^{+},p_g^{-}\in\mathbb{R}^{d}$. We score query feature $z_i$ by its difference in cosine similarity to the two prototypes:

\begin{equation}
q_i^{(g)}
=
\operatorname{sim}_{\cos}(z_i,p_g^{+})
-
\operatorname{sim}_{\cos}(z_i,p_g^{-}),
\qquad
\operatorname{sim}_{\cos}(a,b)
=
\frac{a^{\top}b}{\|a\|_2\|b\|_2}.
\label{eq:zs-score}
\end{equation}

The support sets are constructed for each granularity $g$: globally, per anomaly category, and within the same video. Per-category and within-video prototypes are constructed with knowledge of ground truth snippet labels, whereas the global prototype is based on video labels only. For the within-video prototype, the evaluated query snippet is excluded from its support set through a leave-one-out procedure.~\Cref{fig:three-granularities-fig} (bottom) illustrates the three-granularity evaluation under zero-shot prototypes. We evaluate the resulting scores $q_i^{(g)}$ using Equations ~\eqref{eq:auc-global}--\eqref{eq:auc-within-video}.

\subsection{Resolving WSVAD Model Comparisons}
\label{sec:paired-bootstrap}

The multi-granularity analysis asks whether the pooled AUC reflects the intended within-video localization ability, both for the detector's final scores and for prototype readouts of its learned representations. A separate question is whether the margins observed at any granularity are large enough to support the resulting model ranking. For such question, it is worth distinguishing two properties. First, AUC is a statistically well-understood metric that is relatively stable under repeated sampling and invariant to class prevalence in its pairwise interpretation~\cite{li2024area}. This stability concerns how \emph{precisely} AUC estimates a \emph{fixed} comparison population. Second, the statistical stability of AUC as a metric does not imply that a finite benchmark has the sufficient resolution for AUC to reliably distinguish models with scores differing by only fractions of a point. In most VAD benchmarks, where the test set is finite in terms of unique video scenes, the variation of AUC estimates under different video resamples can be several magnitudes wider than the margins between state of the art estimates.

\begin{figure}
    \centering
    \includegraphics[width=\linewidth]{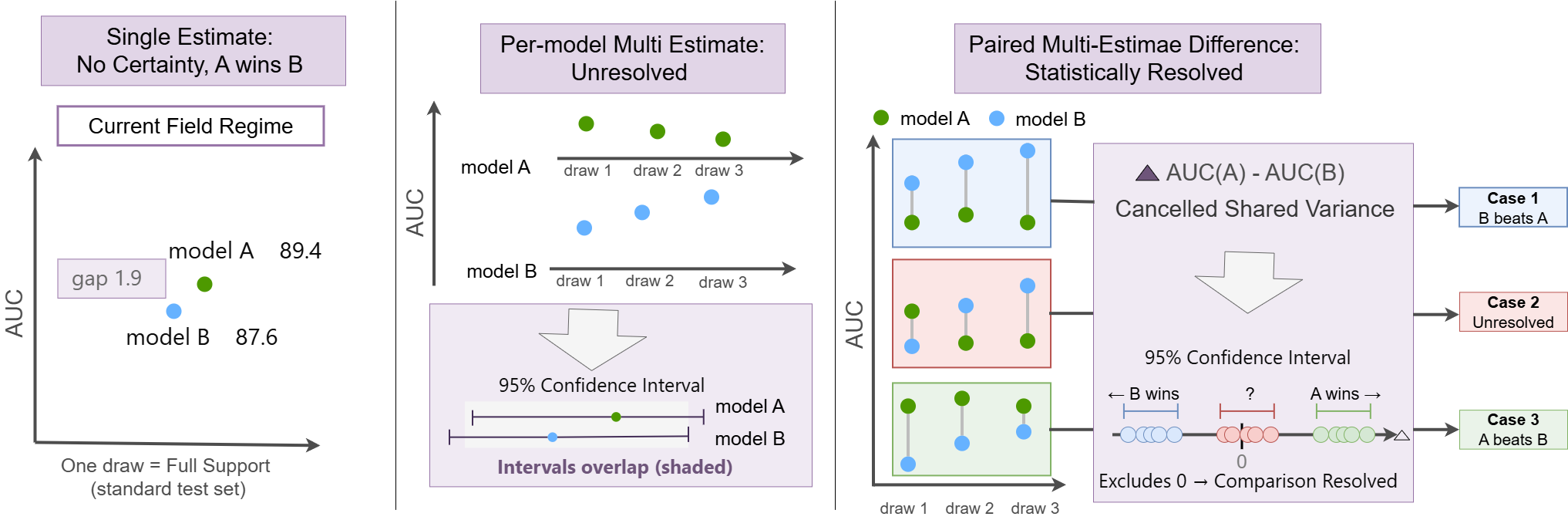}
    \caption{\textbf{AUC ranking resolution.} Left: a single AUC per model. Middle: each model’s AUC distribution across video resamples yields overlapping 95\% intervals, leaving comparison unresolved. Right: the paired AUC difference resolves the ranking as it removes the shared effect on the video resamples. }
\label{fig:resolution}
\end{figure}

We quantify the AUC uncertainty by repeatedly drawing evaluation videos with replacement, retaining all frames of each selected video, and recomputing AUC on each of the $N$ resampled sets. The $2.5^{th}$ and $97.5^{th}$ percentiles of the resulting AUC distribution form a $95\%$ confidence interval. Such an interval quantifies uncertainty in one model's AUC estimate, but when intervals across models are overlapping, it does not differentiate a model from another. Therefore, model ranking requires the confidence interval of the paired AUC \emph{difference}, computed from identical video resamples for both models.~\Cref{fig:resolution} contrasts these readings on an illustrative model pair. Our paired video bootstrap and AUC difference comparison is a nonparametric counterpart to the DeLong test~\cite{delong1988comparing}, which compares two correlated AUCs analytically via U-statistic theory, under the samples independence assumption. Since frames within a video are strongly correlated, we resample at the video level to preserve this dependence structure.

Mathematically, let $\mathcal{V}$ denote the set of evaluation videos with $n=|\mathcal{V}|$. For each resample $r=1,\ldots,B$, where $B$ is the total number of resamples, we draw $n$ videos from $\mathcal{V}$ with replacement forming the video multi-set $I_r$. All frames of each selected video are retained, and every model is evaluated on the same $I_r$. Let $\mathrm{AUC}_g^m(I_r)$ denote the AUC of model $m$ at granularity $g$ computed on replicate $I_r$. For a model pair $(A,B)$, the paired AUC difference $D_{AB}^{(r)}$ is
\begin{equation}
D_{AB,g}^{(r)}
=
\mathrm{AUC}_g^A(I_r)
-
\mathrm{AUC}_g^B(I_r).
\label{eq:paired-diff}
\end{equation}
We form a $95\%$ confidence interval for $D_{AB,g}$ by taking the central 95\% of its empirical distribution $\{D_{AB,g}^{(r)}\}_{r=1}^{B}$ across resamples. That is, the $2.5^{th}$ and $97.5^{th}$ percentiles denoted as $Q_{0.025}$ and $Q_{0.975}$:

\begin{equation}
\mathrm{CI}_{95\%}(D_{AB,g})
=
\left[
Q_{0.025}(\mathcal{D}_{AB,g}),
Q_{0.975}(\mathcal{D}_{AB,g})
\right].
\label{eq:paired-ci}
\end{equation}
A model pair comparison is considered resolved at granularity $g$ when $0\notin\mathrm{CI}_{95\%}(D_{AB,g})$. For model set $\mathcal{M}$, we summarize granularity $g$ by the number of resolved unordered model pairs:
\begin{equation}
C_g
=
\sum_{\{A,B\}\subset\mathcal{M}}
\mathbf{1}\!\left[
0\notin\mathrm{CI}_{95\%}(D_{AB,g})
\right],
\qquad
0\le C_g\le\binom{|\mathcal{M}|}{2}.
\label{eq:resolution-count}
\end{equation}
Thus, $C_g$ measures how many model comparisons AUC can statistically settle at granularity $g$.

\section{WSVAD Scene Reliance Audit}
\label{sec:scene-control}

\begin{wrapfigure}{r}{0.65\textwidth}
\centering
\begin{minipage}[htb!]{\linewidth}
 \vspace*{-0.8cm}
    \centering
    \includegraphics[width=\linewidth]{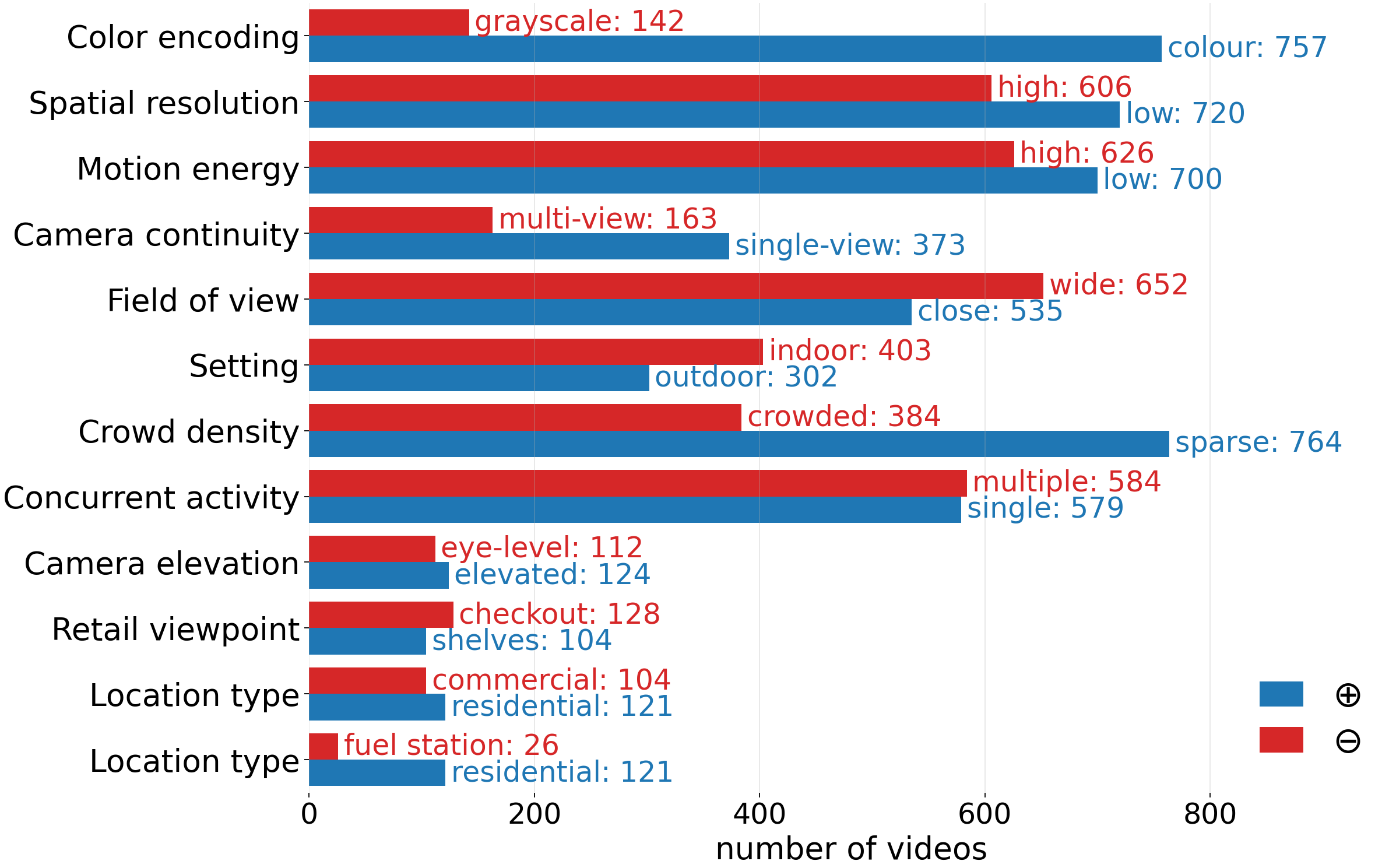}
    \caption{\textbf{Support of the scene-factor audit.} }
    \label{fig:scene-factor-support}
\end{minipage}
\vspace*{-0.85cm} 
\end{wrapfigure}
The three-granularity breakdown and the AUC resolution show whether a model benefits from cross-video comparisons and whether its ranking remains consistent. However, for any resolved model comparison, none of these evaluations quantify whether the performance gain is genuine (from the use of anomaly cues), or confound (from exploiting the recording properties and semantic context that correlate with the anomaly events). We tackle this by assessing model bias when it scores predefined scene factors in normal video footage.


\paragraph{Scene factor annoation.} We annotate the full UCF-Crime videos~\cite{sultani2018real} with twelve binary scene factors covering acquisition conditions, viewpoint, environment, location, motion, and content composition. For each factor, a video is assigned to one of two poles ($\oplus$ or $\ominus$), or marked \emph{ambiguous}. Ambiguous videos are excluded only from that factor, thus, each factor has its own evaluation support (\Cref{fig:scene-factor-support}). Full annotation definitions and protocol details are provided in Supp.~\ref{sup:scene-reliance-audit}.

\paragraph{Measuring model bias.} We compare model performance on normal frames only from different scene factor groups. Since each factor has two opposite poles, we quantify a model bias by its AUC ranking separation between two scene poles. We call this \emph{bias-AUC}. As the anomaly label is constant across the comparison, any departure from $0.5$ reflects a response to the scene factor rather than discrimination between anomalous and normal content. Writing $\bar{s}_v^0$ for the mean score over the normal frames of video $v$, the video-level \emph{bias-AUC} for factor $j$ is \begin{equation}
b_j
=
\Pr\!\left(
\bar{s}_{v_{\oplus}}^0 >
\bar{s}_{v_{\ominus}}^0
\right)
+
\tfrac{1}{2}
\Pr\!\left(
\bar{s}_{v_{\oplus}}^0 =
\bar{s}_{v_{\ominus}}^0
\right),
\quad
v_{\oplus}\sim\mathcal{V}_j^{\oplus},
\quad
v_{\ominus}\sim\mathcal{V}_j^{\ominus}.
\label{eq:bias-auc}
\end{equation}
where $\mathcal{V}_j^{\oplus}$ and $\mathcal{V}_j^{\ominus}$ denote the
sets of videos assigned to the two opposing poles of scene factor $j$. The \emph{bias-AUC} is obtained as the ROC-AUC with pole $\oplus$ treated as the positive class. Values above $0.5$ indicate higher normal-frame scores at the positive pole, and below $0.5$ indicates higher scores at the negative pole. We compute $d_j^{\mathrm{bias}} = |b_j - 0.5|$ when comparing strength irrespective of sign.

\paragraph{Uncertainty and practical significance.} We resample videos with replacement, stratified within each factor pole so that both are still represented. We keep each sampled video's frames together to preserve temporal dependence, and recompute \Cref{eq:bias-auc} over $B$ replicates. We report the percentile $95\%$ interval and 
\begin{equation}
p_{\mathrm{bias}}^{(j)} =
\frac{1}{B}\sum_{r=1}^{B}
\mathbf{1}\!\left[\left|b_j^{(r)} - 0.5\right| > 0.05\right],
\label{eq:p-bias}
\end{equation}
which is the fraction of replicates falling outside a practically-unbiased band of $\pm 0.05$. This distinguishes a factor with a small but consistent effect from one whose observed effect varies depending on which videos happened to be selected. We further comment on the effect of unequal pole sizes in Supp.~\ref{sup:unequal-poles}.

\begin{table}[t]
\centering
\caption{%
  \textbf{Models assessed in this study.} Feature backbones correspond to the original architectures adopted by the authors. AUC values are taken from the reported results, except for Universal MIL, where $\dagger$ indicates our implementation. Probe location is the representation preceding the final classification layer in each model.}
\label{tab:model_zoo}
\begingroup
\small
\setlength{\tabcolsep}{4pt}
\resizebox{\linewidth}{!}{%
\begin{tabular}{l l l c l c}
\toprule
\textbf{Model} & \textbf{Venue} & \textbf{Probe location} & \textbf{Probe\ dim} & \textbf{Feature backbone} & AUC (\%) \\
\midrule
Universal MIL$\dagger$~\cite{trvad}
  & CVPR'18
  & Pre-classifier hidden layer (pre-$fc_2$)
  & 512
  & I3D
  & 79.68 \\
UR-DMU~\cite{urdmu}
  & AAAI'23
  & Pre-classifier encoder ($\mathbf{x}_{\text{pre-cls}}$)
  & 1024
  & I3D
  & 86.97 \\
BN-WVAD~\cite{zhou2024batchnorm}
  & T-CSVT'24
  & Pre-classifier post-attention ($\bar{\mathbf{x}}_{\text{att}}$)
  & 512
  & I3D
  & 87.24 \\
PEL4VAD~\cite{pel4vad}
  & TIP'24
  & Pre-classifier visual projection ($\mathbf{x}_v$)
  & 704
  & I3D
  & 86.76 \\
$\pi$-VAD~\cite{pivad}
  & CVPR'25
  & Pre-classifier encoder ($\mathbf{x}_{\text{pre-cls}}$)
  & 1024
  & I3D
  & 90.33 \\
GS-MoE~\cite{gs-moe}
  & ICCV'25
  & Pre-classifier encoder ($\mathbf{x}_{\text{pre-cls}}$)
  & 1024
  & I3D
  & 91.58 \\
TEVAD~\cite{tevad}
  & CVPR'23
  & Pre-classifier fusion ($\mathbf{x}_v \oplus \mathbf{x}_t$)
  & 2816
  & I3D + SwinBERT
  & 84.92 \\
  \midrule
Universal MIL$\dagger$~\cite{trvad}
  & CVPR'18
  & Pre-classifier hidden layer (pre-$fc_2$)
  & 512
  & VideoMAE\,v2 (ViT-g/14)
  & 85.48 \\
SST-WSVADL~\cite{sst-wsvadl}
  & ECCV'26
  & Pre-classifier encoder ($\mathbf{x}_{\text{pre-cls}}$)
  & 1024
  & VideoMAE\,v2 (ViT-g/14)
  & 88.52 \\
  \midrule
Universal MIL$\dagger$~\cite{trvad}
  & CVPR'18
  & Pre-classifier hidden layer (pre-$fc_2$)
  & 512
  & CLIP (ViT-L/14)
  & 84.47 \\
VadCLIP~\cite{wu2024vadclip}
  & AAAI'24
  & Pre-classifier ($\mathbf{v} + \mathrm{MLP}_2(\mathbf{v})$)
  & 512
  & CLIP (ViT-B/16)
  & 88.02 \\
DSANet~\cite{dsanet}
  & AAAI'26
  & Pre-classifier ($\mathbf{v} + \mathrm{MLP}_2(\mathbf{v})$)
  & 512
  & CLIP (ViT-B/16)
  & 89.40 \\
TrCLIP-VAD~\cite{trvad}
  & Neural Net.'26
  & Pre-classifier ($\mathbf{f} + \mathrm{MLP}_2(\mathbf{f})$)
  & 1280
  & CLIP (ViT-L/14) + text
  & 88.59 \\
\bottomrule
\end{tabular}%
}
\endgroup
\end{table}


\section{Experimental Results}
\label{sec:results}

\subsection{Dataset and Implementation Details}

The headline analysis in this work emphasizes comparisons across recent top-tier WSVAD methods with different backbone families (Table~\ref{tab:model_zoo}). We adopt the official UCF-Crime~\cite{sultani2018real} dataset, with frame-level temporal annotations from~\cite{sst-wsvadl}, and the public I3D, VideoMAEv2, and CLIP features from prior works~\cite{urdmu,zhou2024batchnorm,pivad,tevad,wu2024vadclip,sst-wsvadl}. Snippet size is set to 16 in all experiments. For CLIP-based variants on frame-level features, we substitute the center frame feature for the snippet. Scores are computed from the public pre-trained checkpoints, except for a few models we re-train (marked in~\Cref{tab:model_zoo,tab:zs_vs_score_auc_granularity}). Prototype-based AUC is computed on probe features from the frozen VAD models. Score-based AUC is calculated on the models' predictions. Prob-AUC is computed based on soft multi-annotator labels~\cite{liu2025rethinking}. Resampling experiments use $B=1000$ video-level replicates, and all AUCs are reported on the official UCF-Crime 290-test support. The scene-factor audit (\Cref{sec:scene-control}) has a larger support, including all train and test videos for adequate scene diversity. The audit is descriptive rather than a performance estimate. No model is trained, selected, or calibrated on these annotations.

\begin{table*}[htbp!]
  \caption{\textbf{Prototype-based AUC vs.\ Score-based AUC vs.\ Prob-AUC at three granularities.}
  Backbone-only rows (VideoMAEv2, I3D, CLIP) have no Score-based AUC or Prob-AUC entry. $^\dagger$Indicates our re-training. Model ranks are marked as \colorbox{rankone}{Top-1$^{st}$}, \colorbox{ranktwo}{Top-2$^{nd}$}, and \colorbox{rankthree}{Top-3$^{rd}$} . }
  \label{tab:zs_vs_score_auc_granularity}
  \centering
  \setlength{\tabcolsep}{3.0pt}
  \resizebox{\linewidth}{!}{%
  \begin{tabular}{lccc|ccc|ccc}
    \toprule
    & \multicolumn{3}{c|}{\textbf{Prototype-based AUC}}
    & \multicolumn{3}{c|}{\textbf{Score-based AUC}}
    & \multicolumn{3}{c}{\textbf{Prob-AUC}} \\
    \cmidrule(lr){2-4}\cmidrule(lr){5-7}\cmidrule(lr){8-10}
    \textbf{Feature / model}
    & \textbf{Global}
    & \textbf{Per-cat.}
    & \textbf{Within-video}
    & \textbf{Global}
    & \textbf{Per-cat.}
    & \textbf{Within-video}
    & \textbf{Global}
    & \textbf{Per-cat.}
    & \textbf{Within-video} \\
    
    \midrule
    \cellcolor{tan} VideoMAEv2
      & \cellcolor{tan} 83.87 & \cellcolor{tan} \corr{70.63} & \cellcolor{tan} \corr{\rktwo{93.88}}
      & \cellcolor{tan} \textemdash & \cellcolor{tan} \textemdash & \cellcolor{tan} \textemdash
      & \cellcolor{tan} \textemdash & \cellcolor{tan} \textemdash & \cellcolor{tan} \textemdash \\
    UR-DMU$\dagger$~\cite{urdmu}
      & \rkone{89.58} & \corr{74.84} & \corr{76.05}
      & 88.08 & 76.50 & 75.83
      & 89.51 & 76.80 & \rktwo{79.40} \\
    PEL4VAD$\dagger$~\cite{pel4vad}
      & \corr{49.55} & \corr{49.19} & \corr{48.72}
      & 87.56 & 73.78 & 76.80
      & 89.26 & 73.12 & 76.77 \\
    BN-WVAD$\dagger$~\cite{zhou2024batchnorm}
      & 84.75 & \corr{74.10} & \corr{84.24}
      & 86.76 & 70.74 & 70.44
      & 88.76 & 71.13 & 74.10 \\
    SST-WSVADL~\cite{sst-wsvadl}
      & \corr{86.92} & \corr{\rkthree{75.45}} & \corr{78.96}
      & 88.52 & \rktwo{78.11} & 75.75
      & 89.48 & \rktwo{77.88} & 78.67 \\
    GS-MoE$\dagger$~\cite{gs-moe}
      & \corr{86.01} & \corr{\rktwo{77.59}} & \corr{77.08}
      & 87.95 & 75.99 & 76.62
      & 89.37 & \rkthree{77.72} & \rkthree{79.36} \\
    Pi-VAD$\dagger$~\cite{pivad}
      & \corr{86.72} & \corr{73.43} & \corr{82.77}
      & 87.99 & \rkthree{77.58} & \rkthree{77.83}
      & \rkthree{89.57} & 75.68 & 76.99 \\
    \midrule
    \cellcolor{lightred} I3D
      & \cellcolor{lightred} \corr{72.79} & \cellcolor{lightred} \corr{65.75} & \cellcolor{lightred} \corr{\rkthree{93.33}}
      & \cellcolor{lightred} \textemdash & \cellcolor{lightred} \textemdash & \cellcolor{lightred} \textemdash
      & \cellcolor{lightred} \textemdash & \cellcolor{lightred} \textemdash & \cellcolor{lightred} \textemdash \\
    UR-DMU~\cite{urdmu}
      & \corr{80.97} & \corr{70.13} & \corr{73.46}
      & 86.97 & 74.22 & 74.13
      & 86.03 & 74.67 & 73.89 \\
    PEL4VAD~\cite{pel4vad}
      & \corr{77.88} & \corr{70.21} & \corr{90.14}
      & 86.76 & 73.22 & 75.42
      & 86.87 & 71.54 & 73.07 \\
    BN-WVAD$\dagger$~\cite{zhou2024batchnorm}
      & \corr{80.56} & \corr{65.69} & \corr{78.15}
      & 84.01 & 65.37 & 72.10
      & 85.65 & 66.25 & 70.79 \\
    SST-WSVADL~\cite{sst-wsvadl}
      & \corr{81.86} & \corr{67.27} & \corr{73.17}
      & 86.50 & 74.46 & 73.73
      & 85.76 & 74.85 & 73.33 \\
    GS-MoE~\cite{gs-moe}
      & \corr{\rktwo{89.24}} & \corr{\rkone{84.84}} & \corr{85.91}
      & \rkone{91.58} & \rkone{83.86} & \rkone{86.79}
      & 89.26 & \rkone{80.25} & \rkone{82.59} \\
    Pi-VAD~\cite{pivad}
      & \corr{86.72} & \corr{73.43} & \corr{82.77}
      & \rktwo{90.33} & 76.88 & \rktwo{79.66}
      & 87.31 & 72.68 & 77.70 \\
    TEVAD~\cite{tevad}
      & \corr{83.92} & \corr{71.81} & \corr{92.46}
      & 84.92 & 66.53 & 70.79
      & 84.61 & 66.25 & 70.79 \\
    \midrule
    \cellcolor{blue} CLIP
      & \cellcolor{blue} \corr{78.18} & \cellcolor{blue} \corr{67.31} & \cellcolor{blue} \corr{90.09}
      & \cellcolor{blue} \textemdash & \cellcolor{blue} \textemdash & \cellcolor{blue} \textemdash
      & \cellcolor{blue} \textemdash & \cellcolor{blue} \textemdash & \cellcolor{blue} \textemdash \\
    VadCLIP~\cite{wu2024vadclip}
      & \corr{83.46} & \corr{71.27} & \corr{92.82}
      & 87.79 & 70.30 & 67.64
      & 89.47 & 71.39 & 74.75 \\
    DSANet~\cite{dsanet}
      & \corr{\rkthree{87.94}} & \corr{72.19} & \corr{92.66}
      & \rkthree{89.44} & 75.74 & 69.81
      & \rkone{90.10} & 75.92 & 75.61 \\
    TrCLIP-VAD~\cite{trvad}
      & \corr{87.23} & \corr{73.56} & \corr{\rkone{95.81}}
      & 88.59 & 72.01 & 73.37
      & \rktwo{89.66} & 70.18 & 75.36 \\
    \bottomrule

  \end{tabular}}
\end{table*}

\subsection{Anomaly Signal in Learned Representations}
\label{sec:representation-results}

\Cref{tab:zs_vs_score_auc_granularity} (left-most column group) reports prototype-based ZS-AUC at the three granularities, computed without the model's classification head. At the video granularity, the anomaly direction is built per video, leaving the query snippet out when constructing the normal and abnormal prototypes. Since ground truth frame labels are used to construct the prototypes, we treat AUC as an oracle diagnostic ceiling on accessible signal, not as achievable detection performance.

The frozen backbones exhibit a dominant within-video anomaly representation. VideoMAEv2 and I3D achieve a within-video ZS-AUC of \corr{$93.33-93.88$}, while CLIP reaches \corr{$90.09$}. Training WSVAD models on these features does not consistently strengthen this signal. Compared to VideoMAEv2, within-video ZS-AUC decreases to \corr{$76.05$} for UR-DMU, \corr{$84.24$} for BN-WVAD, and \corr{$78.96$} for SST-WSVADL. This indicates that the learned representations attenuate the linearly accessible anomaly structure already present in the frozen backbone. In contrast, CLIP-based models exhibit a different behavior, improving the oracle ZS-AUC over the frozen CLIP backbone. These models retain or even strengthen a locally accessible anomaly direction. This suggests that the observed behavior is not a universal consequence of WSVAD training and cannot yet be attributed to a single design choice.

Across backbone families, per-category ZS-AUC sits well above chance but below the global readout. The majority of feature sets fall between $65.69$ and $84.84$, against global ZS-AUC values of $72.79-89.58$. A category-shared anomaly direction is therefore linearly recoverable, where anomalous snippets of one video align with the anomaly-vs-normal contrast of other videos in the same category. However, the shared anomaly direction remains consistently weaker than the global prototype, which may additionally exploit cross-video context that correlates with the anomaly label. The one exception is PEL4VAD (VideoMAEv2), whose prototype readout is at chance at every granularity ($49.55$, $49.19$, $48.72$). The model's anomaly information seems to be realized only by its head (within-video score based AUC is $76.80$), while not aligning with any class-mean direction. This makes PEL4VAD (VideoMAEv2) a genuine representational outlier rather than one point in a trend.

\subsection{Representation Signal vs. Detection Performance}
\label{sec:prediction-results}

This section reads each model's own anomaly scores at the three granularities, reporting what the trained head delivers rather than what dominates in the representations. 

\paragraph{Representation signal does not predict localization.} We observe that within-video score-based and prototype-based AUC disagree, and the disagreement reverses the model ranking (\Cref{tab:zs_vs_score_auc_granularity}, \emph{Within-video} prototype and score columns). The five models with the highest prototype AUC (\corr{$90.14-95.81$}) deliver the \emph{lowest} within-video score AUC ($67.64-75.42$). Conversely, the models with lower prototype AUC (\corr{$48.72-77.08$}) deliver higher score AUC (\corr{$73.73-76.80$}). The extremes make this observation concrete; PEL4VAD (VideoMAEv2) localizes at $76.80$ from a prototype AUC of \corr{$48.72$} (at chance), whereas VadCLIP reaches a \corr{$92.82$} prototype ceiling but only $67.64$ within-video score AUC. Thus, the readability of a video-specific anomaly direction from the representation appears to be decoupled from the effectiveness of the scoring head in ranking anomalous against normal snippets in that video. This dissociation is not an artifact of the original temporal annotations. Under soft multi-annotator labels~\cite{liu2025rethinking}(\Cref{tab:zs_vs_score_auc_granularity}, Prob-AUC), absolute values and several ranks shift, yet the mismatch with Prototype-AUC persists. The effect is therefore robust to annotation variation on the score side. 

\paragraph{Pooled AUC cannot separate the methods it ranks.} 
Across all granularities in~\Cref{fig:granularities-spread}, the per-model bootstrap intervals overlap substantially. At the pooled granularity, every model's interval contains every other model's AUC estimate. As overlapping intervals do not settle the models ranking, we therefore test the paired difference directly by computing \Cref{eq:paired-diff,eq:paired-ci,eq:resolution-count} on all model pairs and counting the resolved comparisons. \Cref{tab:resolved-pairs} presents the resolution counts, and \Cref{tab:resolved-pair-details} details the resolved pairs. 

\begin{wraptable}{r}{5.2cm}
\centering
\begin{minipage}[htb!]{\linewidth}
 \vspace*{-1.2cm}
\caption{\textbf{Model pairs resolved under each granularity.} Spread is max\,-\,min of point estimates in the subset.}
\label{tab:resolved-pairs}
\centering
\small
\setlength{\tabcolsep}{4.5pt}
\resizebox{\linewidth}{!}{
\begin{tabular}{@{}llrcc@{}}
\toprule
Subset & Granularity & Pairs resolved & & Spread (pts) \\
\midrule
\multirow{3}{*}{All (16 models)}
  & Global       & 34 / 120 & & 7.64 \\
  & Per-category & 55 / 120 & & 19.56 \\
  & Within-video & 53 / 120 & & 19.15 \\
\midrule
\multirow{3}{*}{VideoMAEv2 (6)}
  & Global       & \textbf{0} / 15 & & 1.75 \\
  & Per-category & 5 / 15  & & 7.37 \\
  & Within-video & 5 / 15  & & 7.38 \\
\midrule
\multirow{3}{*}{I3D (7)}
  & Global       & 9 / 21  & & 7.64 \\
  & Per-category & 14 / 21 & & 19.56 \\
  & Within-video & 13 / 21 & & 16.00 \\
\midrule
\multirow{3}{*}{CLIP (3)}
  & Global       & \textbf{0} / 3 & & 1.65 \\
  & Per-category & 1 / 3   & & 5.44 \\
  & Within-video & 1 / 3   & & 5.73 \\
\bottomrule
\end{tabular}
}
\end{minipage}
\vspace*{-0.8cm}
\end{wraptable}

Analyzing Tables~\ref{tab:resolved-pairs} and \ref{tab:resolved-pair-details} jointly, the overall resolved counts seem to overstate the AUC's discriminative power. Among $34$ resolved pairs across the model zoo, $25$ are the easy cross-backbone comparisons, e.g. an I3D-based model against a CLIP-based one. Such comparisons are rarely informative, since a new method claims the state of the art by outperforming prior methods on the \emph{same} features. Restricted to the same-backbone comparisons, global AUC resolves $0$ of $15$ VideoMAEv2-based pairs and $0$ of $3$ CLIP-based pairs, while the within-video readout resolves $5$ and $1$ of the same pairs. Even though the I3D family appears to be the exception ($9$ of $21$ resolved), its resolved pairs are consistently centred on two models. The first is GS-MoE (I3D), which sits $3.5$ points above the rest of the family, and the second is BN-WVAD (I3D), which sits $2.4$ points further below. Among the five competitive I3D models, global AUC resolves zero pairs. Across all three families on UCF-Crime, pooled AUC seems to detect outliers separated by several points and resolves nothing at the margins the field publishes.

Pooled AUC also reorders models relative to within-video localization. DSANet holds the highest pooled AUC among CLIP models ($89.44$) but the lowest within-video AUC of the three ($69.81$). Models ranked at or below DSANet by pooled AUC, including VideoMAEv2-based Pi-VAD, PEL4VAD, GS-MoE, UR-DMU, and SST-WSVADL, localize better within video by up to $7.99+$ (resolved). The reversal holds across backbone families, thus, it reflects what the two protocols reward, not a backbone advantage.

This bears directly on how improvements are claimed. DSANet is proposed as a refinement of VadCLIP and reports gains across every setting~\cite{dsanet}. While the two are separated per-category ($+4.80$, resolved), the global and within video differences are unresolved, and both models are outperformed by the same set of simpler detectors (\Cref{tab:resolved-pair-details}). 
From observations in \Cref{tab:resolved-pair-details}, no pair separated by a margin $0.4-0.5$ AUC resolves in any backbone family, yet recent methods on UCF-Crime claim improvements within such margins~\cite{dsanet, wu2024vadclip, pel4vad, sst-wsvadl}. 

\begin{figure}[t!]
    \centering
    \begin{subfigure}[t]{0.73\linewidth}
        \centering
        \includegraphics[width=\linewidth]{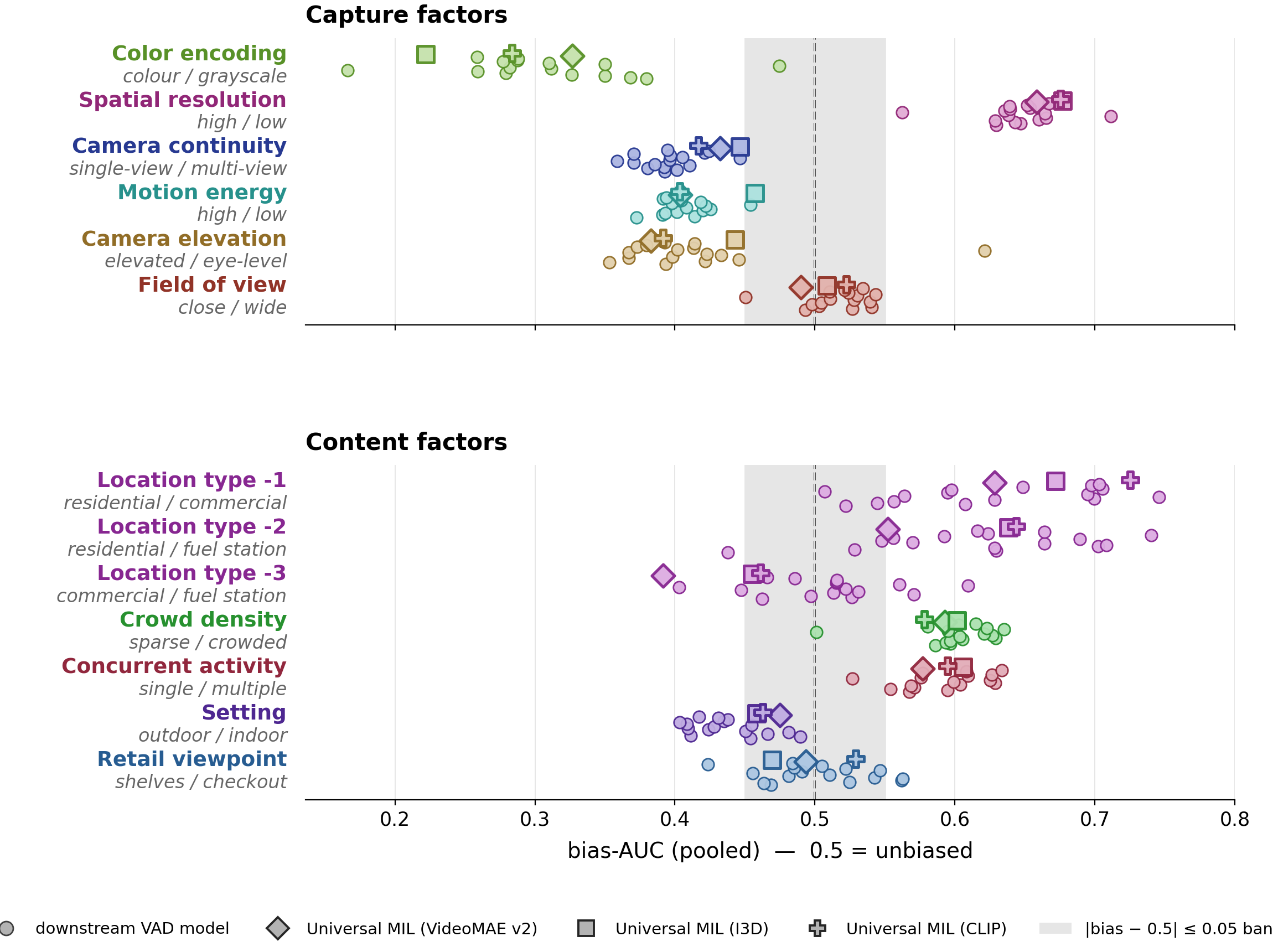}
        \caption{Per-factor \emph{bias-AUC}, one marker per score source.}
        \label{fig:scene-bias-a}
    \end{subfigure}\hfill
    \begin{subfigure}[t]{0.85\linewidth}
        \centering
        \includegraphics[width=\linewidth]{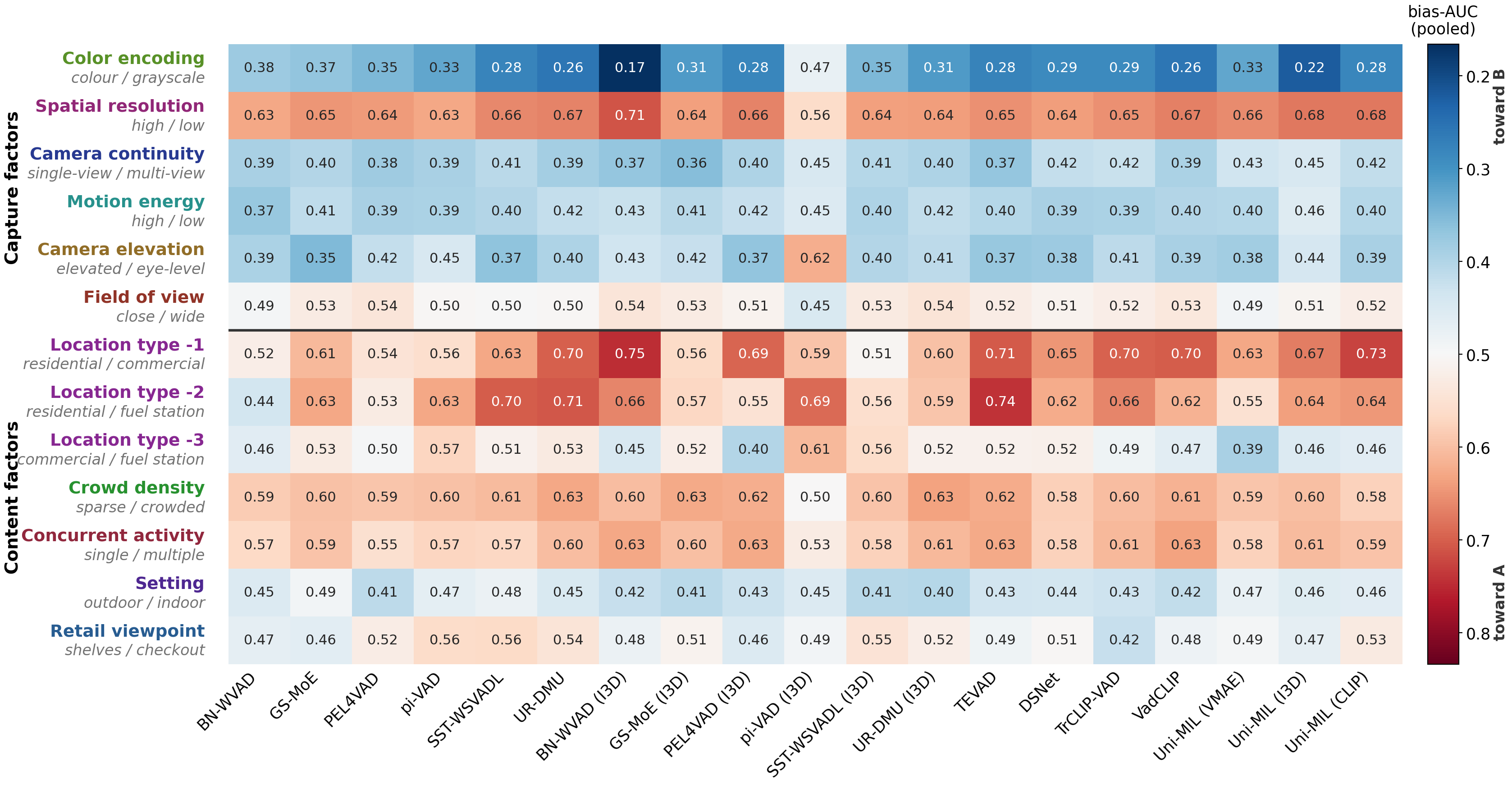}
        \caption{Full model $\times$ factor matrix.}
        \label{fig:scene-bias-b}
    \end{subfigure}
    \caption{\textbf{Scene reliance audit on normal frames only.}}
    \label{fig:scene-bias}
\end{figure}

\subsection{Scene Reliance of Anomaly Scores}
\label{sec:scene-bias}

When anomaly direction is not captured by the class-mean probe of learned representations, as shown in previous sections, the final scores may rely on anomaly-relevant evidence or on recording properties that co-vary with it. We directly evaluate the scene bias, quantified as the deviation of the \emph{bias-AUC} from the null band $p_{\mathrm{bias}} \pm 0.05$ across the twelve labeled scene factors (\Cref{fig:scene-bias}). 

\paragraph{Bias to Recording Properties.}
Two acquisition factors are displaced in the same direction for every score sources. Colour encoding falls below $0.5$ throughout ($0.17$
to $0.47$) and spatial resolution above it ($0.56$ to $0.71$); resolution
reaches $p_{\mathrm{bias}} = 1.000$ for every model. This unanimity spans all backbone families, architectures, and training objectives. Camera continuity, motion energy, and camera elevation follow the same pattern but at smaller magnitude. Field of view is the one acquisition factor with no bias response, where every model sits inside the practically-unbiased band ($0.45$--$0.54$).

\paragraph{Bias to semantic properties.}
Scene response is not confined to acquisition properties. Location type, crowd density, and concurrent activity displace most models above $0.55$, while outdoor-indoor setting sits consistently below $0.5$. These semantic and content biases are real but smaller in magnitude than the acquisition biases, and on most factors the models agree in direction. Only the location-type factors show disagreement in direction across models and backbones.

\paragraph{Backbone and head contributions.}
The bias response does not seem to originate in any trained detector. Universal MIL, a minimal head trained separately on VideoMAEv2, I3D, and CLIP features, shows similar directions on the same factors as the full models (\Cref{fig:scene-bias-b}). The bias magnitude shifts with the feature space more than with the model, therefore models that share a backbone cluster together. For example, the I3D columns show strong color-encoding response for GS-MoE ($0.17$) as well as for the minimal head ($0.22$). Thus, the backbone seems to set the regime, while model architectures and objectives shift values few points around it.

\paragraph{Impact of scene bias on anomaly detection.} A scene-conditioned decomposition of AUC  on anomaly-vs-normal frames shows that the measured bias rarely inflates the anomaly frames scores, except for the color encoding factor. The bias response is present in both anomaly and normal frame scores, thus, it mostly cancels in with frames pooling. See discussion in Supp. \ref{sup:scene-conditioned-auc} and \ref{sup:results-scene-reliance-audit}.

\section{Conclusion}
\label{sec:conclusion}

We asked what pooled frame-level AUC measures on WSVAD benchmarks, such as UCF-Crime, by reading the same anomaly scores under progressively stricter pairing rules and auditing what they respond to. Three key findings follow. First, our experiments on UCF-Crime show that the pooled AUC evaluation protocol lacks the discriminative power to support the rankings built on it. Second, the learned representation by WSVAD models and their predictions are decoupled. The models with the most dominant within-video anomaly structure may localize worst, whereas the strongest within-video detectors may produce representations that no class-mean probe can read. Pooled AUC registers neither of these patterns. Third, every considered model separates normal footage by recording and semantic properties, such as resolution, colour encoding, and location, which carry no anomaly signal by construction.

We do not indict AUC as a statistic, since what fails mostly is the population of comparisons constructed for the pooled protocol. Dominated by frame pairs from different recordings, pooled AUC can reward separability without learning any anomaly concept. The correction is inexpensive and can be derived from the available predictions produced in previous studies. Model comparisons can be resolved by: (1) reporting within-video and per-category AUC alongside the pooled number, and (2) testing claimed improvements as paired differences on shared video resamples. We release our annotations and protocol to make this the path of least resistance.


\putbib[main]
\end{bibunit}

\input{supp}

\end{document}

%% file: supp.tex
\begin{bibunit}[splncs04]
\clearpage
\setcounter{page}{1}


\begingroup
\scriptsize        
\setlength{\parskip}{1pt}

\begin{center}
\large Storyline for \textit{Auditing Frame-Level AUC in Weakly Supervised Video Anomaly Detection (WSVAD): Granularity, Resolution, and Scene Bias}
\end{center}

\paragraph{1. \textbf{Why interesting?}}
\begin{enumerate}[label=\alph*,leftmargin=1.4em,itemsep=1pt]
\item {WSVAD} temporally localizes the {frames} that belong to hazardous incidents or criminal behaviors in long surveillance video.
\item In practice, decisions are made on {one camera stream}: an {anomalous} frame's score must outrank the normal frames' scores in that same stream.
\item Standard benchmarks, such as UCF-Crime~\cite{sultani2018real}, provide only {video-level} binary labels (anomaly/normal) over videos from diverse recordings and scenes; models trained on these labels must localize at the {frame-level}.
\item This encourages {shortcuts}: learning scene or recording properties that co-vary with anomalies rather than the anomalies themselves.
\item The evaluation protocol neither distinguishes nor flags this failure mode, yet it is still used to report new state of the art.
\end{enumerate}

\paragraph{2. \textbf{How done now?}}
\begin{enumerate}[label=\alph*,leftmargin=1.4em,itemsep=1pt]
\item Some work targets spatial (pixel-level) localization rather than temporal detection, but frame-level spatial annotations are scarce.
\item Other work shifts to {VLM}-based models that detect anomalies from contextual descriptions; the confounded protocol is still adopted, and does not separate these methods or establish significance in their reported AUC.
\item Refinements of the protocol report additional metrics alongside the main one.
\end{enumerate}

\paragraph{3. \textbf{What is missing, and So What?}}
\begin{enumerate}[label=\alph*,leftmargin=1.4em,itemsep=1pt]
\item Refining the protocol helps, but most reported metrics share the same problem.
\item The standard metric is {pooled AUC}: ROC-AUC computed over all frame-level predictions pooled across test videos; it reflects only a model's ranking ability.
\item In a {pooled} comparison, the two frames of a {pair} come from different recordings, so {pooled AUC} (and any metric built on the same pairs) can be satisfied by separating recordings, not localizing events. We argue that the {confound} lives in the {pairing}, not the {scoring}.
\end{enumerate}

\paragraph{4. \textbf{Proposed approach (P).}}
\begin{enumerate}[label=\alph*,leftmargin=1.4em,itemsep=1pt]
\item Hold scores fixed; change the {pairing} (which frames are compared), not how they are {scoring}.
\item Read the same scores at three {granularities}: global, per-category, {within-video} (same recording).
\item Probe the {learned representation} directly from pre-trained WSVAD models, to separate what the representation encodes from what the trained head delivers.
\item Test claimed gains as {paired differences} on shared {video resamples} (video-level uncertainty).
\item Audit {scene reliance} directly: score {normal-only} frames by recording or semantic factors.
\end{enumerate}

\paragraph{5. \textbf{Experimental questions.}}
\begin{enumerate}[label=\alph*,leftmargin=1.4em,itemsep=1pt]
\item Does {pooled AUC} reward recording separation over {within-video} localization?\\
c1: same scores, three {granularities}, matched backbones.\\
c2: models rank normal-vs-anomalous well when {pooled} ($\approx$85--90).\\
c3: {within-video} drops sharply; rankings reverse.\\
c4: learned representation probe $\perp$ delivered localization (decoupled).
\item Can the metric resolve the margins the field reports?\\
c1: {paired differences} on shared {video resamples}, per backbone family.\\
c2/c3: 0 pairs resolved at 0.4--0.5 margins; {within-video} resolves several.\\
c4: refining the score ({Prob-AUC}, {soft labels}) leaves ranking unchanged.
\item Does the {scene confound} occur in real data?\\
u1: full UCF-Crime, 12 annotated scene factors.\\
u2: reproduced baselines match published {pooled AUC}.\\
u3: every model separates {normal-only} frames by recording (resolution $p_{\mathrm{bias}}=1.0$).\\
u4: effect is backbone-driven, shared across architectures.
\end{enumerate}

\endgroup

\title{\underline{Supplementary Material} \\Auditing Frame-Level AUC in Weakly Supervised Video Anomaly Detection: Granularity, Resolution, and Scene Bias}

\titlerunning{Auditing Frame-Level AUC in WSVAD} 

\author{Sara Abdulaziz\inst{1}\orcidlink{0009-0002-5204-127X} \and Egor Bondarev \inst{1}\orcidlink{0009-0005-2452-7389} }
\authorrunning{S. Abdulaziz and E. Bondarev}

\institute{Eindhoven University of Technology, 5612 AE Eindhoven, Netherlands\\ \email{s.e.a.m.abdulaziz@tue.nl}\\
\href{https://github.com/Sara-Esam/vad_auc_audit}{GitHub Code}
}

\maketitle

\newcommand{\beginsupplement}{%
    \setcounter{table}{0}
    \renewcommand{\thetable}{S\arabic{table}}
    \setcounter{figure}{0}
    \renewcommand{\thefigure}{S\arabic{figure}}
    \setcounter{section}{0}
    \renewcommand{\thesection}{S\arabic{section}}
}

\beginsupplement


\section{Scene Reliance Audit}\label{sup:scene-reliance-audit}

\subsection{Scene factor Annotation}

We annotate all videos in UCF-Crime according to a set of recording properties and distinguishable scene content (\Cref{tab:scene-factors}). Semantic content factors are chosen empirically for UCF-Crime, based on what the dataset can support. Except for frame resolution and motion level, the rest of the factors were annotated manually. Manual annotation was performed with a purpose-built tool that plays each video and allows selection of multiple factors when present. The annotation was performed by a single annotator, where the video was watched several times at higher speeds for extra verification of the presence of a scene factor. A video is assigned a label only when the criterion holds for the majority of its duration; borderline or mixed cases are marked \emph{ambiguous} and dropped from that factor. Ambiguous video are excluded from the concerning factor comparison but still included for others. Ambiguity was mostly present in factors, such as camera elevation, field of view, and concurrent activities.

\begin{table}[t!]
\centering
\caption{Scene factors annotated on UCF-Crime. Each factor is a binary contrast between two poles ($\oplus$\,/\,$\ominus$); Support is the number of videos assigned to each pole. Level~1 factors describe how the footage was recorded; Level~2 factors describe the scene content. The concurrent-activity factor is reported only as a composition control (its poles correlate with anomaly class by construction). Resolution is labelled by effective sharpness, as UCF-Crime is distributed at a uniform container resolution.}
\label{tab:scene-factors}
\setlength{\tabcolsep}{5pt}
\renewcommand{\arraystretch}{1.15}
\resizebox{\linewidth}{!}{
\begin{tabular}{@{}llp{5.5cm}l@{}}
\toprule
\textbf{Factor} & \textbf{Poles} & \textbf{Criterion} & \textbf{Support } \\
\midrule
\multicolumn{4}{l}{\cellcolor{lightgray} Level 1 Recording properties}\\
\addlinespace[2pt]
Color encoding      & color / grayscale        & RGB colored; IR/nighttime $\to$ grayscale        & 757 / 142 \\
Spatial resolution   & high / low                & effective sharpness: mean Laplacian variance at native resolution, median-split into high/low.            & 606 / 720 \\
Camera continuity    & single-view / multi-view  & one continuous camera vs.\ cuts/zooms/tracking           & 373 / 163 \\
Camera elevation     & elevated / eye-level      & extremely elevated mount vs.\ eye-level or human-height view & 124 / 112 \\
Field of view        & close / wide              & tight framing (single context) vs.\ wide area (multiple context)    & 535 / 652 \\
Motion energy        & high / low                & frame-difference magnitude on normal segments, terciles & 626 / 700 \\
\midrule
\multicolumn{4}{l}{\cellcolor{lightgray}{Level 2 Content composition}}\\
\addlinespace[2pt]
Number of people     & few / many                & $\le 4$ vs.\ $5+$ visible at both normal and abnormal frames; discontinuity is allowed.  & 764 / 384 \\
Setting              & outdoor / indoor          & exterior (street scenes) vs.\ private building interiors (homes, care rooms)                   & 302 / 403 \\
Viewpoint      & shop-shelves / shop-cashair & in-store camera orientation                 & 104 / 128 \\
Location type$^{\dagger}$   & residential / commercial   & house facade/driveway/entrance vs.\ shops/business  & 121 / 104 \\
Location type$^{\dagger}$   & residential / fuel station & shared residential pole; corroboration only & 121 / 26 \\
Location type$^{\dagger}$   & fuel station / commercial & shared poles & 104 / 26 \\
Concurrent activity & single / multiple    & one vs.\ several simultaneous activities; must co-align with anomaly when applicable    & 579 / 584 \\
\bottomrule
\end{tabular}
}
\\[3pt]

\begin{minipage}{\linewidth}
{\fontsize{6}{2}\selectfont
$^{\dagger}$The two location factors share the positive pole (residential) and are not independent.}
\end{minipage}
\end{table}

\subsection{Scene-conditioned Anomaly AUC}\label{sup:scene-conditioned-auc}
As established in \Cref{sec:scene-control}, \emph{bias-AUC} indicates whether anomaly scores respond to a scene property on normal footage. However, this does not determine whether that response contributes to anomaly discrimination. For example, does a model's anomaly identification come partly from bias to a scene factor? In other words, based on the annotated scene factors, how much of a model's anomaly-vs-normal AUC is a scene-factor ranking? We answer this by recomputing the anomaly-vs-normal AUC with cross-pole comparisons removed, and measuring how much the AUC changes.


\paragraph{Setup.}
Fix a scene factor $j$ with two poles $\oplus$ and $\ominus$ (e.g.\ indoor/outdoor), whose per-pole video supports are $\mathcal{V}_j^{\oplus}$ and $\mathcal{V}_j^{\ominus}$. Any anomalous-normal frame pair entering the pooled AUC is one of four \emph{pair types}: (1) same-pole pairs ($\oplus\oplus$, $\ominus\ominus$), which compare frames sharing the factor; (2) cross-pole pairs ($\oplus\ominus$, $\ominus\oplus$), which compare across the factor. 

\paragraph{Reference AUC.}

The standard (pooled) AUC on factor $j$ is a weighted average of the four pair-type AUCs,
\begin{equation}
AUC_{\mathrm{ref}}
=
w_{\oplus\oplus}\,AUC_{\oplus\oplus}
+
w_{\ominus\ominus}\,AUC_{\ominus\ominus}
+
w_{\oplus\ominus}\,AUC_{\oplus\ominus}
+
w_{\ominus\oplus}\,AUC_{\ominus\oplus}.
\label{eq:auc-ref-poles}
\end{equation}

where $w$ is the fraction of all pooled pairs of type $(X,Y)\in\{\oplus,\ominus\}$, such that $\sum_{X,Y} w_{XY}=1$. This is an exact identity: it regroups the same pooled AUC by pair type.

\paragraph{Scene-conditioned AUC.}
We then keep only same-pole pairs and weight the two poles equally,
\begin{equation}
AUC_{\mathrm{cond}}^{(j)}
=
\tfrac{1}{2}\left(AUC_{\oplus\oplus}+AUC_{\ominus\ominus}\right),
\label{eq:scene-conditioned-auc}
\end{equation}
so that every anomalous frame is ranked only against normal frames sharing its scene pole. The drop from the reference to the conditioned readout,
\begin{equation}
\Delta_{\mathrm{scene}}^{(j)}
=
AUC_{\mathrm{ref}}^{(j)}
-
AUC_{\mathrm{cond}}^{(j)},
\label{eq:scene-control-gap}
\end{equation}
measures how much of the anomaly-vs-normal separation on factor $j$ depended on cross-pole comparisons. Since \Cref{eq:scene-conditioned-auc} drops both cross-pole pairs and re-balances the two poles, $\Delta_{\mathrm{scene}}^{(j)}$ is a protocol contrast rather than an additive decomposition of the reference AUC.

\subsection{Bias-AUC under unequal pole sizes}\label{sup:unequal-poles}
The scene-factor supports in \Cref{sec:scene-control} are often imbalanced, as shown in~\Cref{fig:scene-factor-support} and ~\Cref{tab:scene-factors}. For instance, 757 colored videos vs. 142 grayscale videos. This raises a concern: could a reported \emph{bias-AUC} far from 0.5 merely reflect this imbalance rather than a genuine scene response? We answer this briefly. \emph{bias-AUC} estimates a pairwise ranking probability rather than a proportion. Thus, it is invariant to how many videos fall at each pole~\cite{li2024area}. A model that ranks videos neutrally with respect to the factor obtains \emph{bias-AUC} of $0.5$ regardless of pole sizes, rather than drifting toward the majority prevalence. Imbalance instead widens the video-bootstrap interval (fewer videos in the minority pole give a noisier estimate) without shifting its expected value. 

\section{Experimental Results}

\begin{table*}[t]
  \centering
  \scriptsize
  \setlength{\tabcolsep}{3pt}
  \caption{Scene-control gap $\Delta_{\mathrm{scene}}^{(j)}=\mathrm{AUC}_{\mathrm{ref}}^{(j)}-\mathrm{AUC}_{\mathrm{cond}}^{(j)}$ on the full prediction pool. $\mathrm{AUC}_{\mathrm{cond}}$ equal-weights the two same-pole AUCs. Bold marks $|\Delta_{\mathrm{scene}}|\ge 0.02$. Median $|\Delta|$ across all model-factor cells is $0.004$. $^\dagger$ Location poles are residential and commercial. }
  \label{tab:scene-control-gap}
  \resizebox{\linewidth}{!}{
  \begin{tabular}{@{}lcccccccccccc@{}}
    \toprule
    Factor & UR-DMU~\cite{urdmu} & BN-VAD~\cite{zhou2024batchnorm} & PEL~\cite{pel4vad} & SST~\cite{sst-wsvadl} & $\pi$-VAD~\cite{pivad} & GS-MoE~\cite{gs-moe} & VadCLIP~\cite{wu2024vadclip} & DSANet~\cite{dsanet} & TEVAD~\cite{tevad} & TrCLIP~\cite{trvad} & UniMIL~\cite{sultani2018real} & mean \\
    \midrule
    Color encoding & +0.010 & \textbf{+0.028} & +0.020 & +0.016 & \textbf{+0.022} & \textbf{+0.021} & \textbf{+0.034} & \textbf{+0.023} & \textbf{+0.069} & \textbf{+0.036} & \textbf{+0.021} & +0.027 \\

    Resolution & +0.000 & +0.001 & +0.002 & +0.001 & +0.002 & +0.002 &  +0.002 & +0.003 & +0.004 & +0.003 & +0.002 & +0.002 \\

    Continuity & +0.009 & +0.000 & +0.001 & +0.007 & +0.015 & +0.014 & +0.002 & +0.007 & +0.007 & +0.003 & +0.009 & +0.007 \\
    
    Elevation & -0.004 & +0.004 & -0.003 & -0.001 & +0.000 & +0.001 & +0.007 & +0.012 & +0.001 & +0.002 & +0.012 & +0.004 \\
    
    Field of view & +0.001 & +0.002 & +0.000 & +0.001 & -0.002 & +0.000  & +0.001 & -0.002 & -0.001 & +0.002 & +0.001 & +0.000 \\

    Motion & +0.000 & -0.001 & -0.001 & -0.000 & +0.004 & +0.002 &  +0.002 & +0.004 & +0.002 & +0.003 & +0.001 & +0.001 \\

    Crowd & -0.009 & -0.010 & -0.011 & -0.010 & -0.011 & -0.004  & -0.013 & -0.014 & -0.014 & -0.011 & -0.011 & -0.011 \\

    Setting & -0.002 & -0.001 & -0.002 & -0.003 & -0.002 & +0.001 & -0.004 & -0.002 & -0.003 & -0.002 & +0.000 & -0.002 \\
    
    View point & -0.008 & -0.011 & -0.011 & -0.015 & -0.008 & +0.002 & -0.004 & -0.004 & -0.001 & -0.005 & -0.010 & -0.007 \\
    
    $\#$Activities & +0.015 & +0.007 & -0.000 & +0.005 & +0.007 & +0.004 & \textbf{+0.025} & +0.005 & \textbf{+0.026} & +0.014 & +0.006 & +0.011 \\

    Location$\dagger$ &  -0.010 & +0.001 & -0.004 & -0.010 & +0.002 & +0.001 & -0.001 & +0.003 & +0.002 & +0.006 & -0.009 & -0.003 \\
    \bottomrule
  \end{tabular}}
\end{table*}

\begin{table}[t]
  \centering
  \small
  \setlength{\tabcolsep}{5pt}
  \caption{Mean pair-type AUCs per model, averaged over the four significant scene factors (\texttt{residential\_vs\_gasstation}, \texttt{color\_vs\_grayscale}, \texttt{singleact\_vs\_multiact}, \texttt{few-person\_vs\_crowd}; $|\mathrm{mean}\,\Delta_{\mathrm{scene}}|\ge 0.01$). Pole $\oplus$ ($\ominus$) is the first (second) group in each factor name; $\mathrm{AUC}_{\oplus\ominus}$ ranks anomalous frames from pole $\oplus$ against normal frames from pole $\ominus$.}
  \label{tab:avg-pair-auc-significant}
  \begin{tabular}{@{}lccccc@{}}
    \toprule
    Model & $\mathrm{AUC}_{\ominus\ominus}$ & $\mathrm{AUC}_{\oplus\oplus}$ & $\mathrm{AUC}_{\oplus\ominus}$ & $\mathrm{AUC}_{\ominus\oplus}$ & mean $\Delta$ \\
    \midrule
    UR-DMU~\cite{urdmu} & 0.826 & 0.779 & 0.801 & 0.776 & -0.005 \\
    BN-WVAD~\cite{zhou2024batchnorm} & 0.807 & 0.770 & 0.749 & 0.819 & -0.011 \\
    PEL4VAD~\cite{pel4vad} & 0.815 & 0.791 & 0.776 & 0.837 & -0.007 \\
    SST-WSVADL~\cite{sst-wsvadl} & 0.811 & 0.768 & 0.786 & 0.778 & -0.009 \\
    $\pi$-VAD~\cite{pivad} & 0.779 & 0.731 & 0.755 & 0.751 & -0.005 \\
    GS-MoE~\cite{gs-moe} & 0.813 & 0.805 & 0.807 & 0.815 & +0.003 \\
    VadCLIP~\cite{wu2024vadclip} & 0.762 & 0.746 & 0.750 & 0.745 & +0.004 \\
    DSANet~\cite{dsanet} & 0.792 & 0.770 & 0.763 & 0.793 & -0.003 \\
    TEVAD~\cite{tevad} & 0.741 & 0.697 & 0.762 & 0.679 & +0.003 \\
    TrCLIP-VAD~\cite{trvad} & 0.786 & 0.746 & 0.771 & 0.766 & -0.005 \\
    Universal MIL~\cite{sultani2018real} & 0.785 & 0.751 & 0.756 & 0.779 & -0.002 \\
    \bottomrule
  \end{tabular}
\end{table}

\subsection{Anomaly Representations in WSVAD Models}\label{sup:vad-representations}
\paragraph{\corrpar{Why category supervision does not close the per-category gap.}} 
As reported in \Cref{sec:representation-results} (\cref{tab:zs_vs_score_auc_granularity}), per-category prototype AUC sits $10$-$16$ points below the global readout even for models trained with explicit category supervision (VadCLIP and DSANet). The two models produce $71.27$ and $72.19$ per-category prototype AUC against $83.46$ and $87.94$ global prototype AUC. One might expect category supervision to produce a category direction as strong as the global one. The absence of this outcome seems consistent with design choices in both implementations, where neither model scores frames against the class embeddings directly. Instead, both first inject a visual prompt aggregated from the video's own anomaly-weighted features into each class embedding. Thus, the direction $M$ against which frame $i$ is scored for category $c$ is video-conditioned,
\begin{equation}
  M_{i,c} = \langle x_i,\; t_c(v) \rangle ,
  \qquad
  t_c(v) = t_c + g\big(\mathrm{Norm}(A^{\top}X)\big),
\end{equation}
where $X$ and $A$ are video $v$'s frame features and anomaly scores. The category loss is minimized whenever $t_c(v)$ separates the frames of its own video; it never requires $t_c(v) \approx t_c(v')$ across videos of the same category. Thus, the shared direction is not enforced, while the per-category prototype readout measures precisely the shared component.

\begin{figure}[t!]
    \centering
    \includegraphics[width=\linewidth]{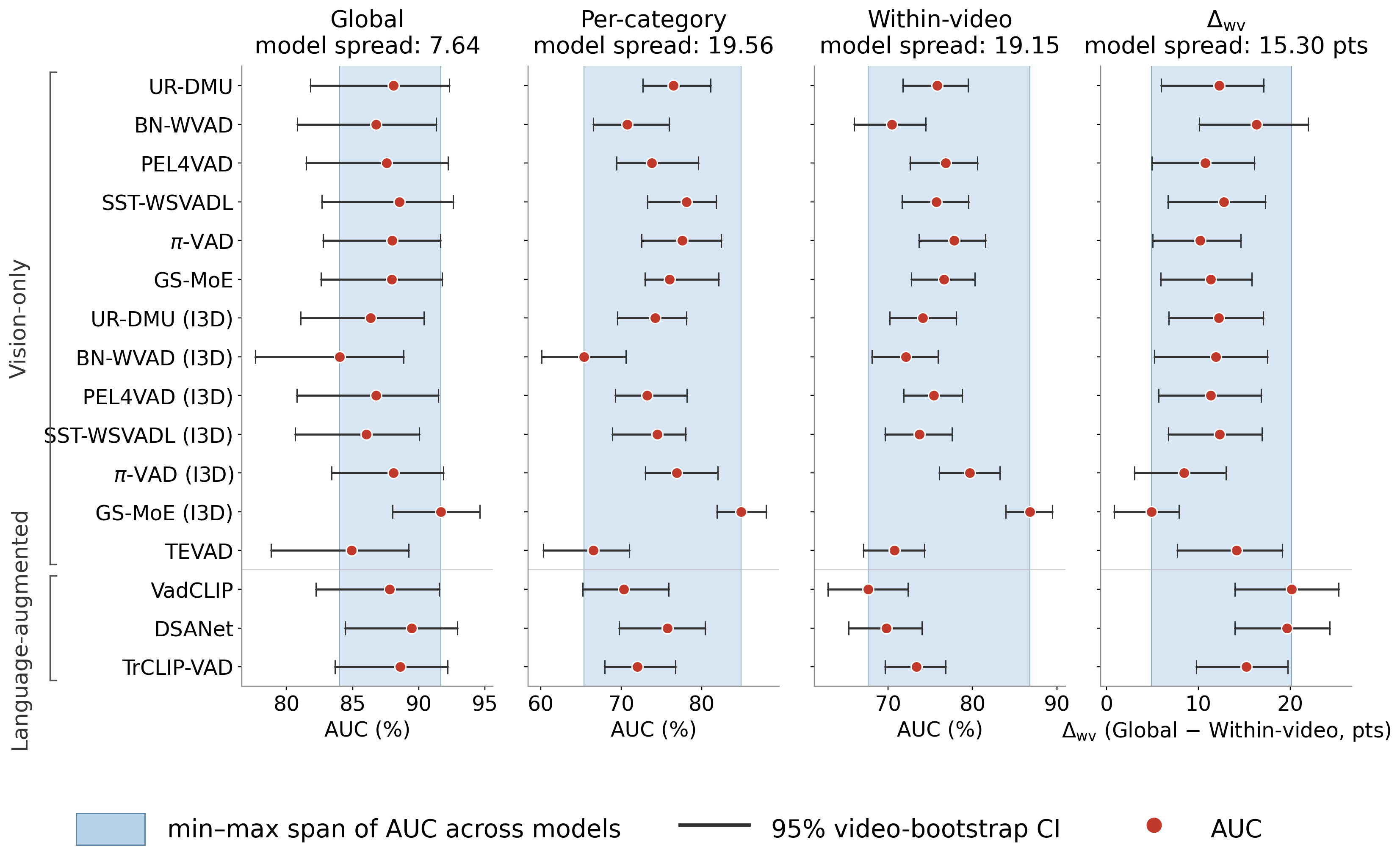}
    \caption{\textbf{Score-based AUC spread across the three granularities.}  The brackets represent the 95\% video-bootstrap CI intervals, and the blue band indicates the region where most of these intervals overlap. $\Delta_{\mathrm{wv}} = \text{Global} - \text{Within-video}$, is the drop from pooled to same-video comparison; larger values indicate a model whose separation relies more on cross-video pairs than on within-video localization}
    \label{fig:granularities-spread}
\end{figure}

\subsection{Scene-Reliance Audit} \label{sup:results-scene-reliance-audit}

\paragraph{Scene bias and anomaly discrimination.}
\Cref{tab:scene-control-gap} reports $\Delta_{\mathrm{scene}}^{(j)}$ for every model-factor cell. The gaps are small for all models ($|\Delta_{\mathrm{scene}}|\le 0.01$). Removing cross-pole comparisons barely changes the anomaly-vs-normal AUC. Therefore, the strong scene responses measured on normal footage (\Cref{sec:scene-bias,fig:scene-bias}) might not translate into inflated detection scores. The bias is likely shifting anomalous and normal frames of a pole together, and canceling once frames are compared within the pole.

\paragraph{Why the cross-pole terms cancel.}
\Cref{tab:avg-pair-auc-significant} shows the four pair-type AUCs, averaged on the top-$4$ factors with the largest $\Delta_{scene}$ gaps. The two cross-pole types move in opposite directions around the same-pole values. In the pooled reference these opposing distortions largely offset, so the net gap stays small even where individual pair types are visibly displaced.

The one factor with a consistent, non-canceling gap is color encoding: $\Delta_{\mathrm{scene}}$ is positive for every model (mean $+0.027$, up to $+0.069$ for TEVAD), meaning the pooled AUC is measurably inflated by color-vs-grayscale comparisons. This validates the audit in \Cref{fig:scene-bias}, where color encoding shows the strongest normal-frame bias. Concurrent activity shows the same effect at smaller magnitude for a subset of models.

Whereas \emph{bias-AUC} establishes that models react to scene factors, $\Delta_{\mathrm{scene}}$ establishes whether that reaction is load-bearing for detection. On UCF-Crime, under the labeled factors, it is mostly not the case. Nonetheless, color encoding is one measurable exception. After all, it should be emphasized that the dominant confound is not cross-\emph{pole} but cross-\emph{video}. Thus, even within one scene pole, the majority of the comparisons still span different recordings. Conditioning on scene factors, therefore, cannot substitute for within-video evaluation.

\begin{table*}[t]
\centering
\caption{\textbf{Resolved pairwise global score-AUC comparisons on full 16 models on UCF-Crime test split.} All $\binom{16}{2}=120$ pairs across VideoMAEv2 (6 models), I3D (7 models), and CLIP (3 models) under the paired video bootstrap of \Cref{eq:paired-diff}. \textbf{Bold} rows are cross-backbone pairs; plain rows are within-backbone.}
\label{tab:resolved-pair-details}
\scriptsize\setlength{\tabcolsep}{3.5pt}
\resizebox{\linewidth}{!}{

\begin{tabular}{@{}llrrrl@{}}
\toprule
Granularity & Winner $\succ$ Loser & Win & Lose & $\Delta$ & $95\%$ CI \\
\midrule
\multirow{34}{*}{Global} 
 & GS-MoE (I3D) $\succ$ BN-WVAD (I3D) & 91.65 & 84.01 & +7.88 & $(+4.77,+11.44)$ \\
 & GS-MoE (I3D) $\succ$ TEVAD & 91.65 & 84.92 & +7.07 & $(+3.68,+11.13)$ \\
 & GS-MoE (I3D) $\succ$ SST-WSVADL (I3D) & 91.65 & 86.02 & +5.80 & $(+3.13,+8.45)$ \\
 & GS-MoE (I3D) $\succ$ UR-DMU (I3D) & 91.65 & 86.35 & +5.45 & $(+2.76,+8.18)$ \\
 & \textbf{DSANet $\succ$ BN-WVAD (I3D)} & \textbf{89.44} & \textbf{84.01} & \textbf{+5.45} & \textbf{$(+2.29,+8.91)$} \\
 & \textbf{GS-MoE (I3D) $\succ$ BN-WVAD} & \textbf{91.65} & \textbf{86.76} & \textbf{+5.05} & \textbf{$(+2.24,+8.43)$} \\
 & GS-MoE (I3D) $\succ$ PEL4VAD (I3D) & 91.65 & 86.76 & +4.85 & $(+2.21,+8.18)$ \\
 & \textbf{TrCLIP-VAD $\succ$ BN-WVAD (I3D)} & \textbf{88.59} & \textbf{84.01} & \textbf{+4.65} & \textbf{$(+1.35,+8.02)$} \\
 & \textbf{DSANet $\succ$ TEVAD} & \textbf{89.44} & \textbf{84.92} & \textbf{+4.63} & \textbf{$(+1.29,+8.02)$} \\
 & \textbf{SST-WSVADL $\succ$ BN-WVAD (I3D)} & \textbf{88.52} & \textbf{84.01} & \textbf{+4.57} & \textbf{$(+1.54,+7.66)$} \\
 & Pi-VAD (I3D) $\succ$ BN-WVAD (I3D) & 88.09 & 84.01 & +4.41 & $(+0.18,+8.50)$ \\
 & \textbf{Pi-VAD $\succ$ BN-WVAD (I3D)} & \textbf{87.99} & \textbf{84.01} & \textbf{+4.18} & \textbf{$(+0.58,+7.91)$} \\
 & \textbf{GS-MoE (I3D) $\succ$ PEL4VAD} & \textbf{91.65} & \textbf{87.56} & \textbf{+4.18} & \textbf{$(+1.25,+7.68)$} \\
 & \textbf{GS-MoE (I3D) $\succ$ VadCLIP} & \textbf{91.65} & \textbf{87.79} & \textbf{+4.13} & \textbf{$(+1.57,+7.19)$} \\
 & \textbf{UR-DMU $\succ$ BN-WVAD (I3D)} & \textbf{88.08} & \textbf{84.01} & \textbf{+4.12} & \textbf{$(+0.49,+7.69)$} \\
 & \textbf{GS-MoE $\succ$ BN-WVAD (I3D)} & \textbf{87.95} & \textbf{84.01} & \textbf{+4.10} & \textbf{$(+0.74,+7.86)$} \\
 & \textbf{TrCLIP-VAD $\succ$ TEVAD} & \textbf{88.59} & \textbf{84.92} & \textbf{+3.83} & \textbf{$(+0.70,+6.94)$} \\
 & \textbf{GS-MoE (I3D) $\succ$ GS-MoE} & \textbf{91.65} & \textbf{87.95} & \textbf{+3.78} & \textbf{$(+1.16,+7.08)$} \\
 & \textbf{GS-MoE (I3D) $\succ$ UR-DMU} & \textbf{91.65} & \textbf{88.08} & \textbf{+3.76} & \textbf{$(+0.64,+7.87)$} \\
 & \textbf{VadCLIP $\succ$ BN-WVAD (I3D)} & \textbf{87.79} & \textbf{84.01} & \textbf{+3.76} & \textbf{$(+0.89,+6.96)$} \\
 & \textbf{SST-WSVADL $\succ$ TEVAD} & \textbf{88.52} & \textbf{84.92} & \textbf{+3.75} & \textbf{$(+0.67,+6.94)$} \\
 & \textbf{PEL4VAD $\succ$ BN-WVAD (I3D)} & \textbf{87.56} & \textbf{84.01} & \textbf{+3.70} & \textbf{$(+0.60,+7.19)$} \\
 & \textbf{GS-MoE (I3D) $\succ$ Pi-VAD} & \textbf{91.65} & \textbf{87.99} & \textbf{+3.70} & \textbf{$(+1.39,+6.94)$} \\
 & GS-MoE (I3D) $\succ$ Pi-VAD (I3D) & 91.65 & 88.09 & +3.47 & $(+0.86,+6.60)$ \\
 & \textbf{DSANet $\succ$ SST-WSVADL (I3D)} & \textbf{89.44} & \textbf{86.02} & \textbf{+3.37} & \textbf{$(+0.45,+6.36)$} \\
 & \textbf{GS-MoE (I3D) $\succ$ SST-WSVADL} & \textbf{91.65} & \textbf{88.52} & \textbf{+3.31} & \textbf{$(+0.72,+6.62)$} \\
 & \textbf{UR-DMU $\succ$ TEVAD} & \textbf{88.08} & \textbf{84.92} & \textbf{+3.31} & \textbf{$(+0.01,+6.82)$} \\
 & \textbf{GS-MoE (I3D) $\succ$ TrCLIP-VAD} & \textbf{91.65} & \textbf{88.59} & \textbf{+3.23} & \textbf{$(+1.09,+5.73)$} \\
 & PEL4VAD (I3D) $\succ$ BN-WVAD (I3D) & 86.76 & 84.01 & +3.04 & $(+0.42,+5.49)$ \\
 & \textbf{VadCLIP $\succ$ TEVAD} & \textbf{87.79} & \textbf{84.92} & \textbf{+2.94} & \textbf{$(+0.18,+5.76)$} \\
 & \textbf{BN-WVAD $\succ$ BN-WVAD (I3D)} & \textbf{86.76} & \textbf{84.01} & \textbf{+2.83} & \textbf{$(+0.04,+5.87)$} \\
 & \textbf{DSANet $\succ$ BN-WVAD} & \textbf{89.44} & \textbf{86.76} & \textbf{+2.61} & \textbf{$(+0.25,+5.51)$} \\
 & \textbf{GS-MoE (I3D) $\succ$ DSANet} & \textbf{91.65} & \textbf{89.44} & \textbf{+2.44} & \textbf{$(+0.22,+5.08)$} \\
 & UR-DMU (I3D) $\succ$ BN-WVAD (I3D) & 86.35 & 84.01 & +2.43 & $(+0.19,+4.89)$ \\
\bottomrule
\end{tabular}
}
\end{table*}

\begin{table}[t!]
\caption{\textbf{Resolved pairwise per-category score-AUC comparisons on full 16 models on UCF-Crime test split.} All $\binom{16}{2}=120$ pairs across VideoMAEv2 (6 models), I3D (7 models), and CLIP (3 models) under the paired video bootstrap of \Cref{eq:paired-diff}. \textbf{Bold} rows are cross-backbone pairs; plain rows are within-backbone.}
\label{tab:resolved-pair-details-1}
\centering\scriptsize\setlength{\tabcolsep}{3.5pt}
\resizebox{\linewidth}{!}{

\begin{tabular}{@{}llrrrl@{}}
\toprule
Granularity & Winner $\succ$ Loser & Win & Lose & $\Delta$ & $95\%$ CI \\
\multirow{55}{*}{Per-category} & GS-MoE (I3D) $\succ$ BN-WVAD (I3D) & 84.92 & 65.37 & +19.38 & $(+15.19,+23.78)$ \\
 & GS-MoE (I3D) $\succ$ TEVAD & 84.92 & 66.53 & +19.26 & $(+13.56,+24.63)$ \\
 & \textbf{GS-MoE (I3D) $\succ$ VadCLIP} & \textbf{84.92} & \textbf{70.30} & \textbf{+14.50} & \textbf{$(+10.34,+19.22)$} \\
 & \textbf{GS-MoE (I3D) $\succ$ BN-WVAD} & \textbf{84.92} & \textbf{70.74} & \textbf{+13.44} & \textbf{$(+9.09,+18.06)$} \\
 & \textbf{GS-MoE (I3D) $\succ$ TrCLIP-VAD} & \textbf{84.92} & \textbf{72.01} & \textbf{+12.66} & \textbf{$(+8.89,+16.45)$} \\
 & \textbf{SST-WSVADL $\succ$ BN-WVAD (I3D)} & \textbf{78.11} & \textbf{65.37} & \textbf{+12.30} & \textbf{$(+6.81,+17.49)$} \\
 & \textbf{SST-WSVADL $\succ$ TEVAD} & \textbf{78.11} & \textbf{66.53} & \textbf{+12.18} & \textbf{$(+5.28,+18.05)$} \\
 & \textbf{Pi-VAD $\succ$ BN-WVAD (I3D)} & \textbf{77.58} & \textbf{65.37} & \textbf{+12.14} & \textbf{$(+5.98,+18.23)$} \\
 & \textbf{Pi-VAD $\succ$ TEVAD} & \textbf{77.58} & \textbf{66.53} & \textbf{+12.01} & \textbf{$(+4.47,+18.83)$} \\
 & \textbf{GS-MoE $\succ$ BN-WVAD (I3D)} & \textbf{75.99} & \textbf{65.37} & \textbf{+11.81} & \textbf{$(+5.66,+17.97)$} \\
 & Pi-VAD (I3D) $\succ$ BN-WVAD (I3D) & 76.88 & 65.37 & +11.80 & $(+6.24,+17.16)$ \\
 & \textbf{GS-MoE $\succ$ TEVAD} & \textbf{75.99} & \textbf{66.53} & \textbf{+11.69} & \textbf{$(+4.60,+18.55)$} \\
 & Pi-VAD (I3D) $\succ$ TEVAD & 76.88 & 66.53 & +11.68 & $(+5.47,+17.36)$ \\
 & \textbf{UR-DMU $\succ$ BN-WVAD (I3D)} & \textbf{76.50} & \textbf{65.37} & \textbf{+11.34} & \textbf{$(+6.14,+16.91)$} \\
 & GS-MoE (I3D) $\succ$ SST-WSVADL (I3D) & 84.92 & 74.46 & +11.30 & $(+7.84,+15.28)$ \\
 & GS-MoE (I3D) $\succ$ PEL4VAD (I3D) & 84.92 & 73.22 & +11.29 & $(+7.63,+15.15)$ \\
 & \textbf{UR-DMU $\succ$ TEVAD} & \textbf{76.50} & \textbf{66.53} & \textbf{+11.22} & \textbf{$(+4.53,+17.15)$} \\
 & GS-MoE (I3D) $\succ$ UR-DMU (I3D) & 84.92 & 74.22 & +11.05 & $(+7.42,+14.89)$ \\
 & \textbf{GS-MoE (I3D) $\succ$ PEL4VAD} & \textbf{84.92} & \textbf{73.78} & \textbf{+10.29} & \textbf{$(+5.85,+15.02)$} \\
 & \textbf{GS-MoE (I3D) $\succ$ DSANet} & \textbf{84.92} & \textbf{75.74} & \textbf{+9.71} & \textbf{$(+5.41,+14.34)$} \\
 & \textbf{DSANet $\succ$ BN-WVAD (I3D)} & \textbf{75.74} & \textbf{65.37} & \textbf{+9.67} & \textbf{$(+3.62,+15.43)$} \\
 & \textbf{DSANet $\succ$ TEVAD} & \textbf{75.74} & \textbf{66.53} & \textbf{+9.55} & \textbf{$(+3.32,+15.77)$} \\
 & \textbf{PEL4VAD $\succ$ BN-WVAD (I3D)} & \textbf{73.78} & \textbf{65.37} & \textbf{+9.09} & \textbf{$(+3.82,+15.17)$} \\
 & \textbf{PEL4VAD $\succ$ TEVAD} & \textbf{73.78} & \textbf{66.53} & \textbf{+8.96} & \textbf{$(+2.04,+15.37)$} \\
 & UR-DMU (I3D) $\succ$ BN-WVAD (I3D) & 74.22 & 65.37 & +8.33 & $(+4.54,+12.08)$ \\
 & UR-DMU (I3D) $\succ$ TEVAD & 74.22 & 66.53 & +8.21 & $(+2.19,+14.24)$ \\
 & PEL4VAD (I3D) $\succ$ BN-WVAD (I3D) & 73.22 & 65.37 & +8.09 & $(+4.17,+12.30)$ \\
 & SST-WSVADL (I3D) $\succ$ BN-WVAD (I3D) & 74.46 & 65.37 & +8.08 & $(+4.01,+12.04)$ \\
 & \textbf{GS-MoE (I3D) $\succ$ UR-DMU} & \textbf{84.92} & \textbf{76.50} & \textbf{+8.04} & \textbf{$(+3.97,+12.40)$} \\
 & PEL4VAD (I3D) $\succ$ TEVAD & 73.22 & 66.53 & +7.97 & $(+2.09,+13.96)$ \\
 & SST-WSVADL (I3D) $\succ$ TEVAD & 74.46 & 66.53 & +7.96 & $(+2.16,+13.36)$ \\
 & GS-MoE (I3D) $\succ$ Pi-VAD (I3D) & 84.92 & 76.88 & +7.58 & $(+3.56,+11.68)$ \\
 & \textbf{GS-MoE (I3D) $\succ$ GS-MoE} & \textbf{84.92} & \textbf{75.99} & \textbf{+7.57} & \textbf{$(+2.88,+12.02)$} \\
 & \textbf{SST-WSVADL $\succ$ VadCLIP} & \textbf{78.11} & \textbf{70.30} & \textbf{+7.43} & \textbf{$(+2.72,+12.25)$} \\
 & \textbf{Pi-VAD $\succ$ VadCLIP} & \textbf{77.58} & \textbf{70.30} & \textbf{+7.26} & \textbf{$(+1.61,+12.89)$} \\
 & \textbf{GS-MoE (I3D) $\succ$ Pi-VAD} & \textbf{84.92} & \textbf{77.58} & \textbf{+7.24} & \textbf{$(+3.05,+11.89)$} \\
 & \textbf{GS-MoE (I3D) $\succ$ SST-WSVADL} & \textbf{84.92} & \textbf{78.11} & \textbf{+7.08} & \textbf{$(+3.33,+11.34)$} \\
 & \textbf{GS-MoE $\succ$ VadCLIP} & \textbf{75.99} & \textbf{70.30} & \textbf{+6.94} & \textbf{$(+0.36,+13.23)$} \\
 & \textbf{Pi-VAD (I3D) $\succ$ VadCLIP} & \textbf{76.88} & \textbf{70.30} & \textbf{+6.93} & \textbf{$(+1.19,+12.91)$} \\
 & \textbf{TrCLIP-VAD $\succ$ BN-WVAD (I3D)} & \textbf{72.01} & \textbf{65.37} & \textbf{+6.72} & \textbf{$(+1.95,+11.68)$} \\
 & \textbf{TrCLIP-VAD $\succ$ TEVAD} & \textbf{72.01} & \textbf{66.53} & \textbf{+6.60} & \textbf{$(+0.99,+12.23)$} \\
 & \textbf{UR-DMU $\succ$ VadCLIP} & \textbf{76.50} & \textbf{70.30} & \textbf{+6.47} & \textbf{$(+1.09,+11.79)$} \\
 & SST-WSVADL $\succ$ BN-WVAD & 78.11 & 70.74 & +6.37 & $(+2.81,+10.05)$ \\
 & Pi-VAD $\succ$ BN-WVAD & 77.58 & 70.74 & +6.20 & $(+0.89,+11.81)$ \\
 & \textbf{BN-WVAD $\succ$ BN-WVAD (I3D)} & \textbf{70.74} & \textbf{65.37} & \textbf{+5.94} & \textbf{$(+0.59,+11.28)$} \\
 & GS-MoE $\succ$ BN-WVAD & 75.99 & 70.74 & +5.87 & $(+1.63,+10.20)$ \\
 & \textbf{Pi-VAD (I3D) $\succ$ BN-WVAD} & \textbf{76.88} & \textbf{70.74} & \textbf{+5.86} & \textbf{$(+0.23,+11.48)$} \\
 & \textbf{SST-WSVADL $\succ$ TrCLIP-VAD} & \textbf{78.11} & \textbf{72.01} & \textbf{+5.58} & \textbf{$(+1.39,+9.77)$} \\
 & \textbf{Pi-VAD $\succ$ TrCLIP-VAD} & \textbf{77.58} & \textbf{72.01} & \textbf{+5.42} & \textbf{$(+0.51,+10.27)$} \\
 & UR-DMU $\succ$ BN-WVAD & 76.50 & 70.74 & +5.41 & $(+1.99,+9.12)$ \\
 & \textbf{Pi-VAD (I3D) $\succ$ TrCLIP-VAD} & \textbf{76.88} & \textbf{72.01} & \textbf{+5.08} & \textbf{$(+0.17,+10.14)$} \\
 & DSANet $\succ$ VadCLIP & 75.74 & 70.30 & +4.80 & $(+1.32,+8.64)$ \\
 & \textbf{UR-DMU $\succ$ TrCLIP-VAD} & \textbf{76.50} & \textbf{72.01} & \textbf{+4.62} & \textbf{$(+0.23,+9.13)$} \\
 & \textbf{SST-WSVADL $\succ$ SST-WSVADL (I3D)} & \textbf{78.11} & \textbf{74.46} & \textbf{+4.22} & \textbf{$(+0.04,+8.26)$} \\
 & SST-WSVADL $\succ$ PEL4VAD & 78.11 & 73.78 & +3.22 & $(+0.43,+6.02)$ \\

\bottomrule
\end{tabular}
}
\end{table}

\begin{table}[t!]
\caption{\textbf{Resolved pairwise within-video score-AUC comparisons on full 16 models on UCF-Crime test split.} All $\binom{16}{2}=120$ pairs across VideoMAEv2 (6 models), I3D (7 models), and CLIP (3 models) under the paired video bootstrap of \Cref{eq:paired-diff}. \textbf{Bold} rows are cross-backbone pairs; plain rows are within-backbone.}
\label{tab:resolved-pair-details-2}
\centering\scriptsize\setlength{\tabcolsep}{3.5pt}
\resizebox{\linewidth}{!}{

\begin{tabular}{@{}llrrrl@{}}
\toprule
Granularity & Winner $\succ$ Loser & Win & Lose & $\Delta$ & $95\%$ CI \\
\midrule
\multirow{53}{*}{Within-video} & \textbf{GS-MoE (I3D) $\succ$ VadCLIP} & \textbf{86.79} & \textbf{67.64} & \textbf{+19.24} & \textbf{$(+14.41,+24.03)$} \\
 & \textbf{GS-MoE (I3D) $\succ$ DSANet} & \textbf{86.79} & \textbf{69.81} & \textbf{+17.05} & \textbf{$(+12.64,+21.40)$} \\
 & \textbf{GS-MoE (I3D) $\succ$ BN-WVAD} & \textbf{86.79} & \textbf{70.44} & \textbf{+16.42} & \textbf{$(+11.76,+21.25)$} \\
 & GS-MoE (I3D) $\succ$ TEVAD & 86.79 & 70.79 & +16.08 & $(+12.21,+20.02)$ \\
 & GS-MoE (I3D) $\succ$ BN-WVAD (I3D) & 86.79 & 72.10 & +14.77 & $(+11.22,+18.34)$ \\
 & \textbf{GS-MoE (I3D) $\succ$ TrCLIP-VAD} & \textbf{86.79} & \textbf{73.37} & \textbf{+13.54} & \textbf{$(+9.62,+17.43)$} \\
 & GS-MoE (I3D) $\succ$ SST-WSVADL (I3D) & 86.79 & 73.73 & +13.09 & $(+9.95,+16.63)$ \\
 & GS-MoE (I3D) $\succ$ UR-DMU (I3D) & 86.79 & 74.13 & +12.64 & $(+9.39,+16.07)$ \\
 & \textbf{Pi-VAD (I3D) $\succ$ VadCLIP} & \textbf{79.66} & \textbf{67.64} & \textbf{+12.14} & \textbf{$(+7.20,+17.06)$} \\
 & GS-MoE (I3D) $\succ$ PEL4VAD (I3D) & 86.79 & 75.42 & +11.46 & $(+7.80,+15.16)$ \\
 & \textbf{GS-MoE (I3D) $\succ$ SST-WSVADL} & \textbf{86.79} & \textbf{75.75} & \textbf{+11.01} & \textbf{$(+7.27,+15.10)$} \\
 & \textbf{GS-MoE (I3D) $\succ$ UR-DMU} & \textbf{86.79} & \textbf{75.83} & \textbf{+10.99} & \textbf{$(+7.32,+14.90)$} \\
 & \textbf{Pi-VAD $\succ$ VadCLIP} & \textbf{77.83} & \textbf{67.64} & \textbf{+10.19} & \textbf{$(+4.52,+15.57)$} \\
 & \textbf{GS-MoE (I3D) $\succ$ GS-MoE} & \textbf{86.79} & \textbf{76.62} & \textbf{+10.11} & \textbf{$(+6.27,+13.88)$} \\
 & \textbf{GS-MoE (I3D) $\succ$ PEL4VAD} & \textbf{86.79} & \textbf{76.80} & \textbf{+10.01} & \textbf{$(+5.99,+14.12)$} \\
 & \textbf{Pi-VAD (I3D) $\succ$ DSANet} & \textbf{79.66} & \textbf{69.81} & \textbf{+9.94} & \textbf{$(+4.99,+14.48)$} \\
 & \textbf{Pi-VAD (I3D) $\succ$ BN-WVAD} & \textbf{79.66} & \textbf{70.44} & \textbf{+9.31} & \textbf{$(+4.09,+14.70)$} \\
 & \textbf{PEL4VAD $\succ$ VadCLIP} & \textbf{76.80} & \textbf{67.64} & \textbf{+9.23} & \textbf{$(+3.69,+14.55)$} \\
 & \textbf{GS-MoE $\succ$ VadCLIP} & \textbf{76.62} & \textbf{67.64} & \textbf{+9.13} & \textbf{$(+4.36,+14.07)$} \\
 & \textbf{GS-MoE (I3D) $\succ$ Pi-VAD} & \textbf{86.79} & \textbf{77.83} & \textbf{+9.05} & \textbf{$(+5.00,+13.18)$} \\
 & Pi-VAD (I3D) $\succ$ TEVAD & 79.66 & 70.79 & +8.97 & $(+4.52,+13.59)$ \\
 & \textbf{UR-DMU $\succ$ VadCLIP} & \textbf{75.83} & \textbf{67.64} & \textbf{+8.25} & \textbf{$(+3.83,+12.98)$} \\
 & \textbf{SST-WSVADL $\succ$ VadCLIP} & \textbf{75.75} & \textbf{67.64} & \textbf{+8.23} & \textbf{$(+3.90,+12.97)$} \\
 & \textbf{Pi-VAD $\succ$ DSANet} & \textbf{77.83} & \textbf{69.81} & \textbf{+7.99} & \textbf{$(+2.77,+12.93)$} \\
 & \textbf{PEL4VAD (I3D) $\succ$ VadCLIP} & \textbf{75.42} & \textbf{67.64} & \textbf{+7.78} & \textbf{$(+2.80,+12.81)$} \\
 & Pi-VAD (I3D) $\succ$ BN-WVAD (I3D) & 79.66 & 72.10 & +7.66 & $(+3.73,+11.51)$ \\
 & Pi-VAD $\succ$ BN-WVAD & 77.83 & 70.44 & +7.36 & $(+3.24,+11.44)$ \\
 & GS-MoE (I3D) $\succ$ Pi-VAD (I3D) & 86.79 & 79.66 & +7.11 & $(+4.63,+9.60)$ \\
 & \textbf{PEL4VAD $\succ$ DSANet} & \textbf{76.80} & \textbf{69.81} & \textbf{+7.04} & \textbf{$(+2.24,+12.00)$} \\
 & \textbf{Pi-VAD $\succ$ TEVAD} & \textbf{77.83} & \textbf{70.79} & \textbf{+7.03} & \textbf{$(+2.08,+12.06)$} \\
 & \textbf{GS-MoE $\succ$ DSANet} & \textbf{76.62} & \textbf{69.81} & \textbf{+6.93} & \textbf{$(+2.85,+11.04)$} \\
 & \textbf{UR-DMU (I3D) $\succ$ VadCLIP} & \textbf{74.13} & \textbf{67.64} & \textbf{+6.60} & \textbf{$(+1.64,+11.81)$} \\
 & \textbf{Pi-VAD (I3D) $\succ$ TrCLIP-VAD} & \textbf{79.66} & \textbf{73.37} & \textbf{+6.43} & \textbf{$(+2.26,+10.38)$} \\
 & PEL4VAD $\succ$ BN-WVAD & 76.80 & 70.44 & +6.41 & $(+3.10,+9.70)$ \\
 & GS-MoE $\succ$ BN-WVAD & 76.62 & 70.44 & +6.30 & $(+2.41,+10.53)$ \\
 & \textbf{SST-WSVADL (I3D) $\succ$ VadCLIP} & \textbf{73.73} & \textbf{67.64} & \textbf{+6.15} & \textbf{$(+1.17,+11.23)$} \\
 & \textbf{PEL4VAD $\succ$ TEVAD} & \textbf{76.80} & \textbf{70.79} & \textbf{+6.07} & \textbf{$(+0.96,+11.28)$} \\
 & \textbf{UR-DMU $\succ$ DSANet} & \textbf{75.83} & \textbf{69.81} & \textbf{+6.06} & \textbf{$(+1.83,+9.93)$} \\
 & \textbf{SST-WSVADL $\succ$ DSANet} & \textbf{75.75} & \textbf{69.81} & \textbf{+6.04} & \textbf{$(+1.94,+10.01)$} \\
 & Pi-VAD (I3D) $\succ$ SST-WSVADL (I3D) & 79.66 & 73.73 & +5.98 & $(+2.22,+9.93)$ \\
 & \textbf{GS-MoE $\succ$ TEVAD} & \textbf{76.62} & \textbf{70.79} & \textbf{+5.97} & \textbf{$(+1.36,+10.90)$} \\
 & \textbf{Pi-VAD $\succ$ BN-WVAD (I3D)} & \textbf{77.83} & \textbf{72.10} & \textbf{+5.72} & \textbf{$(+1.45,+9.84)$} \\
 & TrCLIP-VAD $\succ$ VadCLIP & 73.37 & 67.64 & +5.71 & $(+1.63,+9.79)$ \\
 & \textbf{PEL4VAD (I3D) $\succ$ DSANet} & \textbf{75.42} & \textbf{69.81} & \textbf{+5.58} & \textbf{$(+0.97,+10.12)$} \\
 & Pi-VAD (I3D) $\succ$ UR-DMU (I3D) & 79.66 & 74.13 & +5.54 & $(+1.64,+9.49)$ \\
 & UR-DMU $\succ$ BN-WVAD & 75.83 & 70.44 & +5.43 & $(+2.65,+8.33)$ \\
 & SST-WSVADL $\succ$ BN-WVAD & 75.75 & 70.44 & +5.41 & $(+2.10,+8.74)$ \\
 & \textbf{PEL4VAD (I3D) $\succ$ BN-WVAD} & \textbf{75.42} & \textbf{70.44} & \textbf{+4.95} & \textbf{$(+0.58,+9.44)$} \\
 & \textbf{PEL4VAD $\succ$ BN-WVAD (I3D)} & \textbf{76.80} & \textbf{72.10} & \textbf{+4.76} & \textbf{$(+0.56,+8.86)$} \\
 & \textbf{GS-MoE $\succ$ BN-WVAD (I3D)} & \textbf{76.62} & \textbf{72.10} & \textbf{+4.66} & \textbf{$(+0.35,+9.24)$} \\
 & PEL4VAD (I3D) $\succ$ TEVAD & 75.42 & 70.79 & +4.62 & $(+0.50,+8.80)$ \\
 & Pi-VAD (I3D) $\succ$ PEL4VAD (I3D) & 79.66 & 75.42 & +4.36 & $(+0.68,+8.19)$ \\
 & PEL4VAD (I3D) $\succ$ BN-WVAD (I3D) & 75.42 & 72.10 & +3.31 & $(+0.08,+6.69)$ \\
\bottomrule
\end{tabular}
}
\end{table}


\begin{table*}[t]
  \centering
  \scriptsize
  \caption{Scene pair matrices for all standard UCF scene factors (factor names as in the bias-AUC tables). Shaded rows mark the factor.}
  \label{tab:pair-matrix-all-factors-1}
  \setlength{\tabcolsep}{3pt}
  \resizebox{0.70\linewidth}{!}{
  \begin{tabular}{@{}lccccrrrrc@{}}
    \toprule
    Model / factor & $w_{\mathrm{AA}}$ & $w_{\mathrm{BB}}$ & $w_{\mathrm{AB}}$ & $w_{\mathrm{BA}}$ & $\mathrm{AUC}_{\mathrm{AA}}$ & $\mathrm{AUC}_{\mathrm{BB}}$ & $\mathrm{AUC}_{\mathrm{AB}}$ & $\mathrm{AUC}_{\mathrm{BA}}$ & Gap \\
    \midrule
    \rowcolor{gray!20}\multicolumn{10}{@{}l@{}}{\textbf{\texttt{highway\_vs\_sideroad}}} \\
    UR-DMU~\cite{urdmu} & 0.372 & 0.137 & 0.150 & 0.341 & 0.834 & 0.850 & 0.831 & 0.841 & -0.004 \\
    BN-WVAD~\cite{zhou2024batchnorm} & 0.372 & 0.137 & 0.150 & 0.341 & 0.805 & 0.829 & 0.799 & 0.843 & 0.004 \\
    PEL4VAD~\cite{pel4vad} & 0.372 & 0.137 & 0.150 & 0.341 & 0.833 & 0.857 & 0.830 & 0.850 & -0.003 \\
    SST-WSVADL~\cite{sst-wsvadl} & 0.372 & 0.137 & 0.150 & 0.341 & 0.838 & 0.849 & 0.837 & 0.846 & -0.001 \\
    $\pi$-VAD~\cite{pivad} & 0.372 & 0.137 & 0.150 & 0.341 & 0.845 & 0.837 & 0.831 & 0.843 & 0.000 \\
    GS-MoE~\cite{gs-moe} & 0.276 & 0.225 & 0.237 & 0.262 & 0.913 & 0.838 & 0.827 & 0.918 & 0.001 \\
    STPrompt~\cite{stprompt} & 0.372 & 0.137 & 0.150 & 0.341 & 0.877 & 0.790 & 0.748 & 0.908 & 0.023 \\
    VadCLIP~\cite{wu2024vadclip} & 0.373 & 0.137 & 0.150 & 0.341 & 0.833 & 0.824 & 0.787 & 0.865 & 0.007 \\
    DSANet~\cite{dsanet} & 0.373 & 0.137 & 0.150 & 0.341 & 0.862 & 0.800 & 0.821 & 0.849 & 0.012 \\
    TEVAD~\cite{tevad} & 0.372 & 0.137 & 0.150 & 0.340 & 0.802 & 0.795 & 0.818 & 0.790 & 0.001 \\
    TrCLIP-VAD~\cite{trvad} & 0.373 & 0.137 & 0.150 & 0.341 & 0.843 & 0.851 & 0.839 & 0.858 & 0.002 \\
    Universal MIL & 0.372 & 0.137 & 0.150 & 0.341 & 0.859 & 0.788 & 0.808 & 0.842 & 0.012 \\
    \midrule
    \rowcolor{gray!20}\multicolumn{10}{@{}l@{}}{\textbf{\texttt{closeview\_vs\_wideview}}} \\
    UR-DMU~\cite{urdmu} & 0.207 & 0.293 & 0.290 & 0.209 & 0.846 & 0.788 & 0.847 & 0.790 & 0.001 \\
    BN-WVAD~\cite{zhou2024batchnorm} & 0.207 & 0.293 & 0.290 & 0.209 & 0.834 & 0.774 & 0.852 & 0.762 & 0.002 \\
    PEL4VAD~\cite{pel4vad} & 0.207 & 0.293 & 0.290 & 0.209 & 0.833 & 0.792 & 0.848 & 0.773 & 0.000 \\
    SST-WSVADL~\cite{sst-wsvadl} & 0.207 & 0.293 & 0.290 & 0.209 & 0.831 & 0.778 & 0.835 & 0.779 & 0.001 \\
    $\pi$-VAD~\cite{pivad} & 0.207 & 0.293 & 0.290 & 0.209 & 0.807 & 0.765 & 0.780 & 0.792 & -0.002 \\
    GS-MoE~\cite{gs-moe} & 0.227 & 0.272 & 0.279 & 0.222 & 0.848 & 0.825 & 0.852 & 0.820 & 0.000 \\
    STPrompt~\cite{stprompt} & 0.207 & 0.293 & 0.290 & 0.209 & 0.783 & 0.729 & 0.821 & 0.678 & 0.000 \\
    VadCLIP~\cite{wu2024vadclip} & 0.207 & 0.293 & 0.290 & 0.209 & 0.835 & 0.734 & 0.858 & 0.709 & 0.001 \\
    DSANet~\cite{dsanet} & 0.207 & 0.293 & 0.290 & 0.209 & 0.849 & 0.768 & 0.835 & 0.781 & -0.002 \\
    TEVAD~\cite{tevad} & 0.207 & 0.293 & 0.290 & 0.209 & 0.804 & 0.736 & 0.804 & 0.730 & -0.001 \\
    TrCLIP-VAD~\cite{trvad} & 0.207 & 0.293 & 0.290 & 0.209 & 0.840 & 0.757 & 0.856 & 0.745 & 0.002 \\
    Universal MIL & 0.207 & 0.293 & 0.290 & 0.209 & 0.804 & 0.776 & 0.813 & 0.769 & 0.001 \\
    \midrule
    \rowcolor{gray!20}\multicolumn{10}{@{}l@{}}{\textbf{\texttt{singleview\_vs\_multiview}}} \\
    UR-DMU~\cite{urdmu} & 0.475 & 0.095 & 0.178 & 0.253 & 0.852 & 0.812 & 0.793 & 0.865 & 0.009 \\
    BN-WVAD~\cite{zhou2024batchnorm} & 0.475 & 0.095 & 0.178 & 0.253 & 0.833 & 0.829 & 0.817 & 0.839 & 0.000 \\
    PEL4VAD~\cite{pel4vad} & 0.475 & 0.095 & 0.178 & 0.253 & 0.835 & 0.833 & 0.778 & 0.874 & 0.001 \\
    SST-WSVADL~\cite{sst-wsvadl} & 0.475 & 0.095 & 0.178 & 0.253 & 0.843 & 0.812 & 0.801 & 0.852 & 0.007 \\
    $\pi$-VAD~\cite{pivad} & 0.475 & 0.095 & 0.178 & 0.253 & 0.833 & 0.770 & 0.771 & 0.834 & 0.015 \\
    GS-MoE~\cite{gs-moe} & 0.431 & 0.112 & 0.166 & 0.291 & 0.829 & 0.763 & 0.769 & 0.823 & 0.014 \\
    STPrompt~\cite{stprompt} & 0.475 & 0.095 & 0.178 & 0.253 & 0.783 & 0.731 & 0.673 & 0.820 & 0.011 \\
    VadCLIP~\cite{wu2024vadclip} & 0.475 & 0.095 & 0.178 & 0.253 & 0.767 & 0.779 & 0.683 & 0.852 & 0.002 \\
    DSANet~\cite{dsanet} & 0.475 & 0.095 & 0.178 & 0.253 & 0.823 & 0.793 & 0.766 & 0.843 & 0.007 \\
    TEVAD~\cite{tevad} & 0.475 & 0.095 & 0.178 & 0.253 & 0.792 & 0.758 & 0.695 & 0.834 & 0.007 \\
    TrCLIP-VAD~\cite{trvad} & 0.475 & 0.095 & 0.178 & 0.253 & 0.805 & 0.800 & 0.750 & 0.847 & 0.003 \\
    Universal MIL & 0.475 & 0.095 & 0.178 & 0.253 & 0.838 & 0.797 & 0.797 & 0.836 & 0.009 \\
    \midrule
    \rowcolor{gray!20}\multicolumn{10}{@{}l@{}}{\textbf{\texttt{color\_vs\_grayscale}}} \\
    UR-DMU~\cite{urdmu} & 0.660 & 0.033 & 0.111 & 0.196 & 0.809 & 0.780 & 0.623 & 0.898 & 0.010 \\
    BN-WVAD~\cite{zhou2024batchnorm} & 0.660 & 0.033 & 0.111 & 0.196 & 0.815 & 0.733 & 0.765 & 0.790 & 0.028 \\
    PEL4VAD~\cite{pel4vad} & 0.660 & 0.033 & 0.111 & 0.196 & 0.820 & 0.763 & 0.763 & 0.817 & 0.020 \\
    SST-WSVADL~\cite{sst-wsvadl} & 0.660 & 0.033 & 0.111 & 0.196 & 0.802 & 0.757 & 0.668 & 0.852 & 0.016 \\
    $\pi$-VAD~\cite{pivad} & 0.660 & 0.033 & 0.111 & 0.196 & 0.772 & 0.717 & 0.630 & 0.831 & 0.022 \\
    GS-MoE~\cite{gs-moe} & 0.647 & 0.036 & 0.114 & 0.204 & 0.798 & 0.740 & 0.733 & 0.804 & 0.021 \\
    STPrompt~\cite{stprompt} & 0.660 & 0.033 & 0.111 & 0.196 & 0.739 & 0.669 & 0.437 & 0.867 & 0.024 \\
    VadCLIP~\cite{wu2024vadclip} & 0.660 & 0.033 & 0.111 & 0.196 & 0.785 & 0.690 & 0.585 & 0.846 & 0.034 \\
    DSANet~\cite{dsanet} & 0.660 & 0.033 & 0.111 & 0.196 & 0.798 & 0.734 & 0.658 & 0.842 & 0.023 \\
    TEVAD~\cite{tevad} & 0.660 & 0.033 & 0.111 & 0.196 & 0.782 & 0.593 & 0.580 & 0.799 & 0.069 \\
    TrCLIP-VAD~\cite{trvad} & 0.660 & 0.033 & 0.111 & 0.196 & 0.793 & 0.694 & 0.647 & 0.824 & 0.036 \\
    Universal MIL & 0.660 & 0.033 & 0.111 & 0.196 & 0.777 & 0.717 & 0.672 & 0.802 & 0.021 \\
    \midrule
    \rowcolor{gray!20}\multicolumn{10}{@{}l@{}}{\textbf{\texttt{highresolution\_vs\_lowresolution}}} \\
    UR-DMU~\cite{urdmu} & 0.303 & 0.194 & 0.318 & 0.185 & 0.794 & 0.845 & 0.905 & 0.689 & 0.000 \\
    BN-WVAD~\cite{zhou2024batchnorm} & 0.303 & 0.194 & 0.318 & 0.185 & 0.778 & 0.818 & 0.850 & 0.726 & 0.001 \\
    PEL4VAD~\cite{pel4vad} & 0.303 & 0.194 & 0.318 & 0.185 & 0.779 & 0.841 & 0.879 & 0.724 & 0.002 \\
    SST-WSVADL~\cite{sst-wsvadl} & 0.303 & 0.194 & 0.318 & 0.185 & 0.777 & 0.836 & 0.887 & 0.696 & 0.001 \\
    $\pi$-VAD~\cite{pivad} & 0.303 & 0.194 & 0.318 & 0.185 & 0.736 & 0.825 & 0.837 & 0.719 & 0.002 \\
    GS-MoE~\cite{gs-moe} & 0.310 & 0.187 & 0.325 & 0.178 & 0.780 & 0.868 & 0.885 & 0.755 & 0.002 \\
    STPrompt~\cite{stprompt} & 0.303 & 0.194 & 0.318 & 0.185 & 0.704 & 0.782 & 0.882 & 0.538 & 0.002 \\
    VadCLIP~\cite{wu2024vadclip} & 0.303 & 0.194 & 0.318 & 0.185 & 0.742 & 0.807 & 0.877 & 0.625 & 0.002 \\
    DSANet~\cite{dsanet} & 0.303 & 0.194 & 0.318 & 0.185 & 0.761 & 0.823 & 0.866 & 0.700 & 0.003 \\
    TEVAD~\cite{tevad} & 0.303 & 0.194 & 0.318 & 0.185 & 0.721 & 0.791 & 0.856 & 0.630 & 0.004 \\
    TrCLIP-VAD~\cite{trvad} & 0.303 & 0.194 & 0.318 & 0.185 & 0.755 & 0.817 & 0.871 & 0.671 & 0.003 \\
    Universal MIL & 0.303 & 0.194 & 0.318 & 0.185 & 0.744 & 0.814 & 0.868 & 0.661 & 0.002 \\
    \midrule
        \rowcolor{gray!20}\multicolumn{10}{@{}l@{}}{\textbf{\texttt{outdoor\_vs\_indoor}}} \\
    UR-DMU~\cite{urdmu} & 0.199 & 0.301 & 0.199 & 0.301 & 0.871 & 0.788 & 0.823 & 0.840 & -0.002 \\
    BN-WVAD~\cite{zhou2024batchnorm} & 0.199 & 0.301 & 0.199 & 0.301 & 0.854 & 0.758 & 0.732 & 0.869 & -0.001 \\
    PEL4VAD~\cite{pel4vad} & 0.199 & 0.301 & 0.199 & 0.301 & 0.872 & 0.776 & 0.801 & 0.848 & -0.002 \\
    SST-WSVADL~\cite{sst-wsvadl} & 0.199 & 0.301 & 0.199 & 0.301 & 0.858 & 0.786 & 0.813 & 0.831 & -0.003 \\
    $\pi$-VAD~\cite{pivad} & 0.199 & 0.301 & 0.199 & 0.301 & 0.831 & 0.758 & 0.785 & 0.808 & -0.002 \\
    GS-MoE~\cite{gs-moe} & 0.210 & 0.293 & 0.236 & 0.261 & 0.814 & 0.825 & 0.795 & 0.842 & 0.001 \\
    STPrompt~\cite{stprompt} & 0.199 & 0.301 & 0.199 & 0.301 & 0.794 & 0.722 & 0.742 & 0.776 & -0.001 \\
    VadCLIP~\cite{wu2024vadclip} & 0.199 & 0.301 & 0.199 & 0.301 & 0.814 & 0.729 & 0.751 & 0.785 & -0.004 \\
    DSANet~\cite{dsanet} & 0.199 & 0.301 & 0.199 & 0.301 & 0.838 & 0.767 & 0.793 & 0.814 & -0.002 \\
    TEVAD~\cite{tevad} & 0.199 & 0.301 & 0.199 & 0.301 & 0.825 & 0.706 & 0.755 & 0.784 & -0.003 \\
    TrCLIP-VAD~\cite{trvad} & 0.199 & 0.301 & 0.199 & 0.301 & 0.833 & 0.755 & 0.774 & 0.815 & -0.002 \\
    Universal MIL & 0.199 & 0.301 & 0.199 & 0.301 & 0.838 & 0.767 & 0.768 & 0.837 & 0.000 \\

    \bottomrule
  \end{tabular}}
\end{table*}

\begin{table*}[t]
  \centering
  \scriptsize
  \caption{\Cref{tab:pair-matrix-all-factors-1} continued. Scene pair matrices for all standard UCF scene factors (factor names as in the bias-AUC tables). Shaded rows mark the factor.}
  \label{tab:pair-matrix-all-factors}
  \setlength{\tabcolsep}{3pt}
  \resizebox{0.70\linewidth}{!}{
  \begin{tabular}{@{}lccccrrrrc@{}}
    \toprule
    Model / factor & $w_{\mathrm{AA}}$ & $w_{\mathrm{BB}}$ & $w_{\mathrm{AB}}$ & $w_{\mathrm{BA}}$ & $\mathrm{AUC}_{\mathrm{AA}}$ & $\mathrm{AUC}_{\mathrm{BB}}$ & $\mathrm{AUC}_{\mathrm{AB}}$ & $\mathrm{AUC}_{\mathrm{BA}}$ & Gap \\
        \midrule
    \rowcolor{gray!20}\multicolumn{10}{@{}l@{}}{\textbf{\texttt{few-person\_vs\_crowd}}} \\
    UR-DMU~\cite{urdmu} & 0.441 & 0.103 & 0.307 & 0.149 & 0.778 & 0.864 & 0.850 & 0.798 & -0.009 \\
    BN-WVAD~\cite{zhou2024batchnorm} & 0.441 & 0.103 & 0.307 & 0.149 & 0.767 & 0.847 & 0.819 & 0.805 & -0.010 \\
    PEL4VAD~\cite{pel4vad} & 0.441 & 0.103 & 0.307 & 0.149 & 0.780 & 0.849 & 0.803 & 0.840 & -0.011 \\
    SST-WSVADL~\cite{sst-wsvadl} & 0.441 & 0.103 & 0.307 & 0.149 & 0.769 & 0.853 & 0.824 & 0.809 & -0.010 \\
    $\pi$-VAD~\cite{pivad} & 0.441 & 0.103 & 0.307 & 0.149 & 0.742 & 0.833 & 0.812 & 0.768 & -0.011 \\
    GS-MoE~\cite{gs-moe} & 0.454 & 0.103 & 0.272 & 0.171 & 0.825 & 0.851 & 0.844 & 0.833 & -0.004 \\
    STPrompt~\cite{stprompt} & 0.441 & 0.103 & 0.307 & 0.149 & 0.699 & 0.820 & 0.774 & 0.776 & -0.014 \\
    VadCLIP~\cite{wu2024vadclip} & 0.441 & 0.103 & 0.307 & 0.149 & 0.745 & 0.836 & 0.800 & 0.790 & -0.013 \\
    DSANet~\cite{dsanet} & 0.441 & 0.103 & 0.307 & 0.149 & 0.775 & 0.856 & 0.798 & 0.853 & -0.014 \\
    TEVAD~\cite{tevad} & 0.441 & 0.103 & 0.307 & 0.149 & 0.708 & 0.835 & 0.800 & 0.761 & -0.014 \\
    TrCLIP-VAD~\cite{trvad} & 0.441 & 0.103 & 0.307 & 0.149 & 0.763 & 0.847 & 0.812 & 0.812 & -0.011 \\
    Universal MIL & 0.441 & 0.103 & 0.307 & 0.149 & 0.753 & 0.831 & 0.793 & 0.806 & -0.011 \\
    \midrule
    \rowcolor{gray!20}\multicolumn{10}{@{}l@{}}{\textbf{\texttt{singleact\_vs\_multiact}}} \\
    UR-DMU~\cite{urdmu} & 0.223 & 0.239 & 0.408 & 0.131 & 0.759 & 0.797 & 0.836 & 0.707 & 0.015 \\
    BN-WVAD~\cite{zhou2024batchnorm} & 0.223 & 0.239 & 0.408 & 0.131 & 0.768 & 0.767 & 0.800 & 0.721 & 0.007 \\
    PEL4VAD~\cite{pel4vad} & 0.223 & 0.239 & 0.408 & 0.131 & 0.786 & 0.781 & 0.787 & 0.770 & -0.000 \\
    SST-WSVADL~\cite{sst-wsvadl} & 0.223 & 0.239 & 0.408 & 0.131 & 0.764 & 0.782 & 0.794 & 0.747 & 0.005 \\
    $\pi$-VAD~\cite{pivad} & 0.223 & 0.239 & 0.408 & 0.131 & 0.716 & 0.768 & 0.765 & 0.719 & 0.007 \\
    GS-MoE~\cite{gs-moe} & 0.258 & 0.205 & 0.408 & 0.129 & 0.785 & 0.823 & 0.822 & 0.788 & 0.004 \\
    STPrompt~\cite{stprompt} & 0.223 & 0.239 & 0.408 & 0.131 & 0.684 & 0.746 & 0.780 & 0.636 & 0.017 \\
    VadCLIP~\cite{wu2024vadclip} & 0.223 & 0.239 & 0.408 & 0.130 & 0.731 & 0.724 & 0.827 & 0.610 & 0.025 \\
    DSANet~\cite{dsanet} & 0.223 & 0.239 & 0.408 & 0.130 & 0.771 & 0.771 & 0.797 & 0.731 & 0.005 \\
    TEVAD~\cite{tevad} & 0.223 & 0.239 & 0.408 & 0.130 & 0.686 & 0.741 & 0.809 & 0.606 & 0.026 \\
    TrCLIP-VAD~\cite{trvad} & 0.223 & 0.239 & 0.408 & 0.130 & 0.748 & 0.762 & 0.814 & 0.678 & 0.014 \\
    Universal MIL & 0.223 & 0.239 & 0.408 & 0.131 & 0.731 & 0.783 & 0.777 & 0.737 & 0.006 \\
    \midrule
    \rowcolor{gray!20}\multicolumn{10}{@{}l@{}}{\textbf{\texttt{highmotion\_vs\_lowmotion}}} \\
    UR-DMU~\cite{urdmu} & 0.329 & 0.181 & 0.221 & 0.269 & 0.822 & 0.818 & 0.768 & 0.861 & 0.000 \\
    BN-WVAD~\cite{zhou2024batchnorm} & 0.329 & 0.181 & 0.221 & 0.269 & 0.809 & 0.792 & 0.704 & 0.869 & -0.001 \\
    PEL4VAD~\cite{pel4vad} & 0.329 & 0.181 & 0.221 & 0.269 & 0.818 & 0.809 & 0.726 & 0.879 & -0.001 \\
    SST-WSVADL~\cite{sst-wsvadl} & 0.329 & 0.181 & 0.221 & 0.269 & 0.813 & 0.804 & 0.738 & 0.864 & -0.000 \\
    $\pi$-VAD~\cite{pivad} & 0.329 & 0.181 & 0.221 & 0.269 & 0.803 & 0.753 & 0.721 & 0.826 & 0.004 \\
    GS-MoE~\cite{gs-moe} & 0.301 & 0.202 & 0.217 & 0.280 & 0.840 & 0.808 & 0.754 & 0.880 & 0.002 \\
    STPrompt~\cite{stprompt} & 0.329 & 0.181 & 0.221 & 0.269 & 0.762 & 0.722 & 0.673 & 0.799 & 0.003 \\
    VadCLIP~\cite{wu2024vadclip} & 0.329 & 0.181 & 0.221 & 0.269 & 0.788 & 0.761 & 0.701 & 0.832 & 0.002 \\
    DSANet~\cite{dsanet} & 0.329 & 0.181 & 0.221 & 0.269 & 0.815 & 0.767 & 0.721 & 0.849 & 0.004 \\
    TEVAD~\cite{tevad} & 0.329 & 0.181 & 0.221 & 0.269 & 0.772 & 0.746 & 0.694 & 0.811 & 0.002 \\
    TrCLIP-VAD~\cite{trvad} & 0.329 & 0.181 & 0.221 & 0.269 & 0.806 & 0.764 & 0.726 & 0.835 & 0.003 \\
    Universal MIL & 0.329 & 0.181 & 0.221 & 0.269 & 0.790 & 0.771 & 0.702 & 0.843 & 0.001 \\
    \midrule
    \rowcolor{gray!20}\multicolumn{10}{@{}l@{}}{\textbf{\texttt{shopshelves\_vs\_cashier}}} \\
    UR-DMU~\cite{urdmu} & 0.330 & 0.181 & 0.244 & 0.245 & 0.705 & 0.850 & 0.731 & 0.834 & -0.008 \\
    BN-WVAD~\cite{zhou2024batchnorm} & 0.330 & 0.181 & 0.244 & 0.245 & 0.754 & 0.839 & 0.674 & 0.898 & -0.011 \\
    PEL4VAD~\cite{pel4vad} & 0.330 & 0.181 & 0.244 & 0.245 & 0.709 & 0.853 & 0.740 & 0.820 & -0.011 \\
    SST-WSVADL~\cite{sst-wsvadl} & 0.330 & 0.181 & 0.244 & 0.245 & 0.697 & 0.855 & 0.781 & 0.757 & -0.015 \\
    $\pi$-VAD~\cite{pivad} & 0.330 & 0.181 & 0.244 & 0.245 & 0.672 & 0.811 & 0.728 & 0.764 & -0.008 \\
    GS-MoE~\cite{gs-moe} & 0.227 & 0.271 & 0.209 & 0.293 & 0.830 & 0.857 & 0.807 & 0.873 & 0.002 \\
    STPrompt~\cite{stprompt} & 0.330 & 0.181 & 0.244 & 0.245 & 0.738 & 0.795 & 0.740 & 0.797 & -0.003 \\
    VadCLIP~\cite{wu2024vadclip} & 0.330 & 0.181 & 0.244 & 0.245 & 0.796 & 0.841 & 0.795 & 0.840 & -0.004 \\
    DSANet~\cite{dsanet} & 0.330 & 0.181 & 0.244 & 0.245 & 0.811 & 0.850 & 0.809 & 0.847 & -0.004 \\
    TEVAD~\cite{tevad} & 0.330 & 0.181 & 0.244 & 0.245 & 0.750 & 0.771 & 0.723 & 0.798 & -0.001 \\
    TrCLIP-VAD~\cite{trvad} & 0.330 & 0.181 & 0.244 & 0.245 & 0.809 & 0.837 & 0.772 & 0.861 & -0.005 \\
    Universal MIL & 0.330 & 0.181 & 0.244 & 0.245 & 0.690 & 0.824 & 0.694 & 0.821 & -0.010 \\
    \midrule
    \rowcolor{gray!20}\multicolumn{10}{@{}l@{}}{\textbf{\texttt{residential\_area\_vs\_gasstation}}} \\
    UR-DMU~\cite{urdmu} & 0.805 & 0.010 & 0.107 & 0.078 & 0.771 & 0.864 & 0.894 & 0.699 & -0.038 \\
    BN-WVAD~\cite{zhou2024batchnorm} & 0.805 & 0.010 & 0.107 & 0.078 & 0.730 & 0.880 & 0.614 & 0.963 & -0.067 \\
    PEL4VAD~\cite{pel4vad} & 0.805 & 0.010 & 0.107 & 0.078 & 0.780 & 0.869 & 0.750 & 0.921 & -0.036 \\
    SST-WSVADL~\cite{sst-wsvadl} & 0.805 & 0.010 & 0.107 & 0.078 & 0.736 & 0.851 & 0.860 & 0.706 & -0.046 \\
    $\pi$-VAD~\cite{pivad} & 0.805 & 0.010 & 0.107 & 0.078 & 0.695 & 0.800 & 0.812 & 0.687 & -0.040 \\
    GS-MoE~\cite{gs-moe} & 0.707 & 0.022 & 0.185 & 0.085 & 0.812 & 0.840 & 0.830 & 0.835 & -0.008 \\
    STPrompt~\cite{stprompt} & 0.805 & 0.010 & 0.107 & 0.078 & 0.664 & 0.757 & 0.744 & 0.673 & -0.036 \\
    VadCLIP~\cite{wu2024vadclip} & 0.805 & 0.010 & 0.107 & 0.078 & 0.722 & 0.798 & 0.788 & 0.735 & -0.029 \\
    DSANet~\cite{dsanet} & 0.805 & 0.010 & 0.107 & 0.078 & 0.734 & 0.807 & 0.797 & 0.748 & -0.028 \\
    TEVAD~\cite{tevad} & 0.805 & 0.010 & 0.107 & 0.078 & 0.611 & 0.794 & 0.861 & 0.549 & -0.068 \\
    TrCLIP-VAD~\cite{trvad} & 0.805 & 0.010 & 0.107 & 0.078 & 0.681 & 0.840 & 0.811 & 0.749 & -0.059 \\
    Universal MIL & 0.805 & 0.010 & 0.107 & 0.078 & 0.742 & 0.809 & 0.782 & 0.772 & -0.026 \\
    \midrule
    \rowcolor{gray!20}\multicolumn{10}{@{}l@{}}{\textbf{\texttt{residential\_area\_vs\_commrcial}}} \\
    UR-DMU~\cite{urdmu} & 0.242 & 0.257 & 0.272 & 0.229 & 0.771 & 0.705 & 0.903 & 0.499 & -0.010 \\
    BN-WVAD~\cite{zhou2024batchnorm} & 0.242 & 0.257 & 0.272 & 0.229 & 0.730 & 0.753 & 0.699 & 0.794 & 0.001 \\
    PEL4VAD~\cite{pel4vad} & 0.242 & 0.257 & 0.272 & 0.229 & 0.780 & 0.709 & 0.768 & 0.699 & -0.004 \\
    SST-WSVADL~\cite{sst-wsvadl} & 0.242 & 0.257 & 0.272 & 0.229 & 0.736 & 0.697 & 0.774 & 0.604 & -0.010 \\
    $\pi$-VAD~\cite{pivad} & 0.242 & 0.257 & 0.272 & 0.229 & 0.695 & 0.671 & 0.755 & 0.609 & 0.002 \\
    GS-MoE~\cite{gs-moe} & 0.264 & 0.232 & 0.299 & 0.204 & 0.812 & 0.830 & 0.842 & 0.796 & 0.001 \\
    STPrompt~\cite{stprompt} & 0.242 & 0.257 & 0.272 & 0.229 & 0.664 & 0.738 & 0.886 & 0.422 & -0.013 \\
    VadCLIP~\cite{wu2024vadclip} & 0.242 & 0.257 & 0.272 & 0.229 & 0.722 & 0.795 & 0.857 & 0.633 & -0.001 \\
    DSANet~\cite{dsanet} & 0.242 & 0.257 & 0.272 & 0.229 & 0.734 & 0.810 & 0.821 & 0.725 & 0.003 \\
    TEVAD~\cite{tevad} & 0.242 & 0.257 & 0.272 & 0.229 & 0.611 & 0.750 & 0.816 & 0.524 & 0.002 \\
    TrCLIP-VAD~\cite{trvad} & 0.242 & 0.257 & 0.272 & 0.229 & 0.681 & 0.808 & 0.837 & 0.656 & 0.006 \\
    Universal MIL & 0.242 & 0.257 & 0.272 & 0.229 & 0.742 & 0.690 & 0.799 & 0.578 & -0.009 \\
    \bottomrule
  \end{tabular}}
\end{table*}

\FloatBarrier
\clearpage

\putbib[supp]
\end{bibunit}